\pdfoutput=1

\documentclass[11pt]{article}

\usepackage[final]{acl}

\usepackage{times}
\usepackage{latexsym}
\usepackage{graphicx}
\usepackage{subfigure}
\usepackage{booktabs} 
\usepackage[export]{adjustbox}
\usepackage{graphicx}
\usepackage{hyperref}

\usepackage{longtable}

\usepackage[T1]{fontenc}

\usepackage[utf8]{inputenc}

\usepackage{microtype}

\usepackage{inconsolata}

\usepackage{graphicx}

\title{Automatic Instruction Evolving for Large Language Models}

\author{
  Weihao Zeng, Can Xu, Yingxiu Zhao, Jian-Guang Lou, Weizhu Chen\\
  Microsoft \\
  \texttt{\{v-weihaozeng,caxu,v-yingxizhao,jlou,wzchen\}@microsoft.com} \\
}

\usepackage{amsmath}
\usepackage{amssymb}
\usepackage{mathtools}
\usepackage{amsthm}
\usepackage{algorithm}
\usepackage{multirow}
\usepackage{algorithmic}
\usepackage{amsfonts}       
\usepackage{nicefrac}       
\usepackage{microtype}      
\usepackage{mathtools}
\usepackage{colortbl}
\DeclareMathOperator*{\argmax}{arg\,max} 
\newcommand{\cname}{\emph{Auto Evol-Instruct}}

\begin{document}
\maketitle

\begin{abstract}

Fine-tuning large pre-trained language models with Evol-Instruct has achieved encouraging results across a wide range of tasks. However, designing effective evolving methods for instruction evolution requires substantial human expertise. This paper proposes \cname{}, an end-to-end framework that evolves instruction datasets using large language models without any human effort. The framework automatically analyzes and summarizes suitable evolutionary strategies for the given instruction data and iteratively improves the evolving method based on issues exposed during the instruction evolution process. Our extensive experiments demonstrate that the best method optimized by \cname{} outperforms human-designed methods on various benchmarks, including MT-Bench, AlpacaEval, GSM8K, and HumanEval.
\end{abstract}

\section{Introduction}
\label{introduction}

Fine-tuning large language models (LLMs) to follow detailed instructions is vital to unlocking their power \cite{ouyang2022training,touvron2023llama2}. High-quality datasets, such as ShareGPT \cite{chiang2023vicuna}, OpenAssistant \cite{kopf2023openassistant}, LIMA \cite{zhou2023lima}, have greatly improved the performance of instruction-tuning, promoting the prosperity of LLM alignment. However, annotating instruction following datasets with such quality is hard to scale, and its quality upper limit is also uncontrollable. Researchers \cite{xu2023wizardlm,yu2023metamath,liu2023what} are actively exploring ways to break through the quality upper-bound of existing datasets. Evol-Instruct \cite{xu2023wizardlm} takes the high-quality data as a starting point, and further iteratively refines it using LLMs, improving its complexity and diversity. It has demonstrated superior performance across a broad range of public benchmarks that evaluate diverse capabilities, including instruction following \cite{zheng2023judging, alpaca_eval}, code generation \cite{luo2023wizardcoder, chen2021evaluating}, and mathematical reasoning \cite{luo2023wizardmath, cobbe2021training}.

While Evol-Instruct exhibits outstanding performance, its heavy reliance on heuristic efforts presents notable challenges. Whenever it is used for a completely new task, the methods for execution evolution need to be redesigned. Such a process requires a high level of expertise and considerable costs, hindering its adaptation to a wider spectrum of capabilities. To address these challenges, it needs to automate the Evol-Instruct process, which will encounter the following difficulties: (1) Design evolving methods automatically that make the instructions more complex for a given task (2) To keep the instruction evolution process working properly, the evolving method needs to avoid evolution failure.

 \begin{figure*}[t]
 \centering
\resizebox{0.99\textwidth}{!}{
 \includegraphics[scale=0.5]{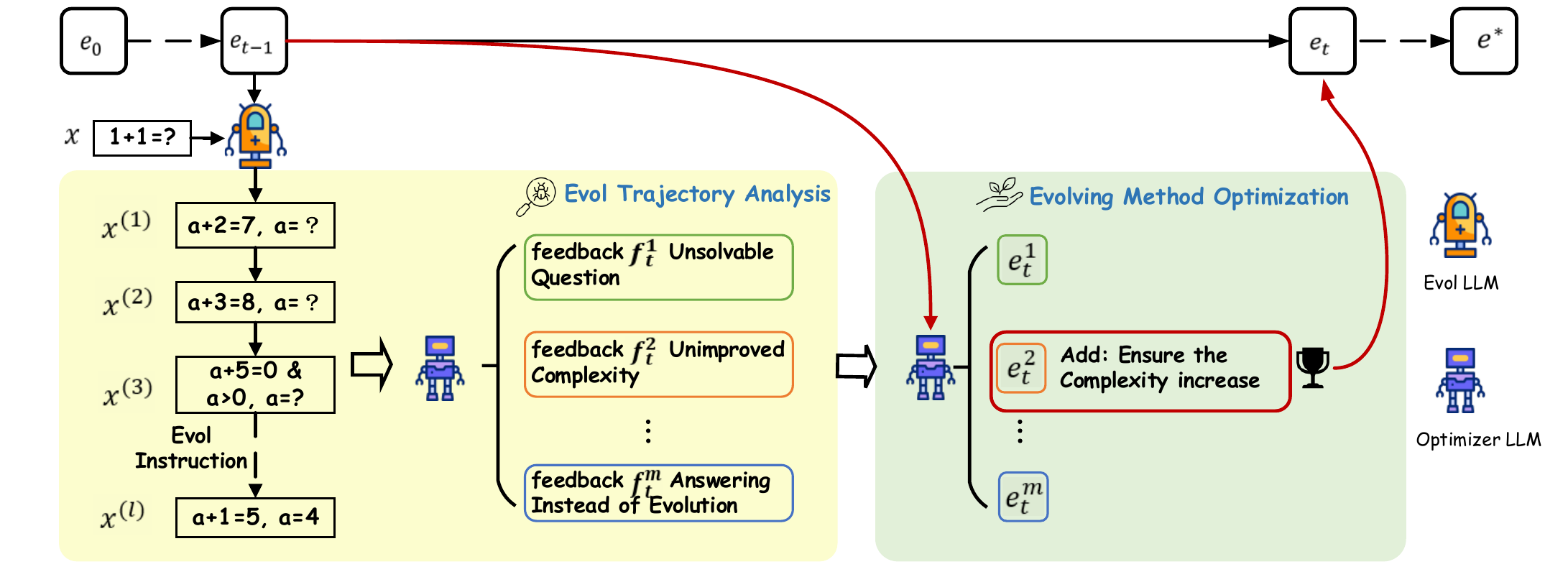}
 }
\vspace{-15pt}
 \caption{Overall architecture of \cname{}. It illustrates the process of optimizing the initial evolving method $e_{0}$ into the optimal evolving method $e^{*}$, which specifically outlines the transition from $e_{t-1}$ to $e_{t}$. The yellow part and green part denote Evol Trajectory Analysis and Evolving Method Optimization respectively. $x^{(1)}$ to $x^{(l)}$ represents the example of evolutionary trajectory obtained by the evol LLM guided by $e_{t-1}$ evolving $x$ for $l$ rounds. The feedback and potential improved evolving methods obtained from $m$ Multiple Optimizations denote $f_{t}^{1}$ to $f_{t}^{m}$ and $e_{t}^{1}$ to $e_{t}^{m}$ respectively.}
 \label{fig:model}
 \vspace{-15pt}

\end{figure*} 

In this paper, we propose \cname{}, an effective approach to utilizing LLMs in designing methods for executing instruction evolution. \cname{} automatically designs evolving methods that make given instruction data more complex, enabling almost cost-free adaptation to different tasks by only changing the input data of the framework. Firstly, to transition from manually-designed evolving rules to automated ones, we begin with a universal initial evolving method. Our initial evolving method is different from the method of Evol Instruct, which requires human experts to specify the rules of evolution. Instead, it can autonomously analyze the input instruction and brainstorm evolution rules suitable for given data. Due to the diversity and complexity of varied instruction datasets, a fixed evolving method can not guarantee the stability and effectiveness of all data evolution. Therefore, we leverage LLM as the optimizer to optimize the initial evolving method iteratively to ensure the lowest failure rate for a given instruction dataset. We refer to the model used for evolution as the evol LLM, and the model used for optimization as the optimizer LLM. This optimization process involves two critical stages: (1) Evol Trajectory Analysis: The optimizer LLM carefully analyzes the potential issues and failures exposed in instruction evolution performed by evol LLM, generating feedback for subsequent optimization. (2) Evolving Method Optimization: The optimizer LLM optimizes the evolving method by addressing these identified issues in feedback. These stages alternate and repeat to progressively develop an effective evolving method using only a subset of the instruction data.  Once the optimal evolving method is identified, it directs the evol LLM to convert the entire instruction dataset into more diverse and complex forms, thus facilitating improved instruction tuning.

Our experiments show that the evolving methods designed by \cname{} outperform the Evol-Instruct methods \cite{xu2023wizardlm, luo2023wizardmath, luo2023wizardcoder} designed by human experts in instruction tuning across various capabilities, including instruction following, mathematical reasoning, and code generation. Using only 10K evolved ShareGPT for fine-tuning Mixtral-8x7B \cite{jiang2024mixtral}, we achieve 8.09 on MT-Bench \cite{zheng2023judging} and 91.4 on AlpacaEval \cite{alpaca_eval}, surpassing GPT-3.5-Turbo and WizardLM-70B, and comparable with Claude-2.0. Using only 7K evolved GSM8K training data for fine-tuning Mixtral-8x7B, we achieve 82.49 on GSM8K, surpassing GPT-3.5-Turbo, WizardMath-70B and MetaMath-70B \cite{yu2023metamath}. Using 20K evolved Code Alpaca to fine-tune DeepSeek-Coder-Base-33B \cite{guo2024deepseek}, we achieve 77.4 on HumanEval, surpassing GPT-3.5-Turbo and WizardCoder-34B.



\section{Background}

\subsection{Evol-Instruct}

Instruction evolution \cite{xu2023wizardlm} involves refining an instruction dataset to boost its complexity and diversity, enhancing instruction tuning effectiveness. This method uses a human-designed evolving method, denoted as $e$, to transform original instruction dataset $X = \{x_1, x_2, \cdots, x_n\}$, where each $x_i$ is an instruction-response pair, into an improved dataset $X_e$. The aim is for $X_e$ to yield superior performance $Q(X_e)$ in a specific capability after instruction tuning, compared to the original dataset's performance $Q(X)$. Essentially, by evolving the instruction dataset and subsequently tuning a model on $X_e$, the model should perform better on the targeted capability than it would using the original dataset.

\subsection{Problem Formulation}

While Evol-Instruct shows excellent performance across many areas, its dependence on high expertise and limited scope restrict its broader use. Our research aims to develop an automated framework that identifies the optimal instruction evolving method, $e^*$, which maximizes performance after instruction tuning:
\begin{equation}
    e^* = \argmax_{e}  Q(X_{e}).
\end{equation}
This framework seeks to find the evolving method $e^*$ that delivers the highest performance $Q(X_e)$ after tuning a model on the evolved dataset $X_e$. By automating this process, we aim to reduce the need for extensive human expertise and expand the application of instruction evolution.


\section{Auto Evol-Instruct}


Unlike Evol-Instruct, \cname{} is a fully automated framework that improves the complexity and quality of instruction data without any human intervention. Its key advancements include: (1) automatically designing evolving methods for instruction evolution, facilitating adaptation to a wide range of tasks and enhancing model capabilities across a broader spectrum; (2) developing evolving methods that surpass those crafted by human experts, while minimizing failures and ensuring successful execution of instruction evolution.


Figure \ref{fig:model} illustrates the process of automating the design of evolving methods in the Auto Evol-Instruct Framework (Section \ref{sec:initial_evol_prompt_design}-\ref{sec:evol_prompt_optimization}).  We also detail specific examples of how the evolving method changes at each step in the Table \ref{tab:case_study}. This framework begins with a carefully designed universal evolving method and a seed instruction dataset $X$ (Section \ref{sec:initial_evol_prompt_design}). It then iteratively optimizes this initial evolving method, $e_{0}$, to obtain the optimal evolving method, $e^{*}$ \footnote{This process uses a subset of the full instruction data, randomly sampling approximately 2,000 entries, to minimize costs associated with developing the evolving method.}. In each optimization step $t$, we randomly sample a mini batch from $X$ and utilize the evol LLM to evolve each instruction in the batch $l$ times. Then the optimizer LLM analyzes the evolutionary trajectory of all instructions in the current batch to identify existing issues and generate feedback (Section \ref{sec:evol_trajectory_ana}). As shown in Figure \ref{fig:model}, the optimizer LLM identifies problems such as “Unimproved Complexity”. The optimizer LLM will make corresponding optimizations to evolving method $e_{t-1}$ to obtain $e_{t}$ based on the feedback. Specifically, the feedback “Unimproved Complexity” will prompt the optimizer LLM to add a constraint “Ensure the Complexity increase” in $e_{t}$. To improve the stability, we execute “analysis optimization” multiple times with sampling decoding in parallel to obtain $m$ optimized evolving methods. Then, we select the method with the lowest evolution failure rate as the final $e_{t}$. The optimization process terminates when the failure rate of instruction evolution no longer decreases, or a maximum number of optimization steps has reached (Section \ref{sec:evol_prompt_optimization}). Once the optimal evolving method is identified, it will be applied to guide the instruction evolution across the entire instruction dataset, resulting in an evolved dataset (Section \ref{sec:evolve_data}).

\subsection{Initial Evolving Method Design}
\label{sec:initial_evol_prompt_design}
\begin{figure*}[t]
 \centering
\resizebox{0.95\textwidth}{!}{
 \includegraphics[scale=0.5]{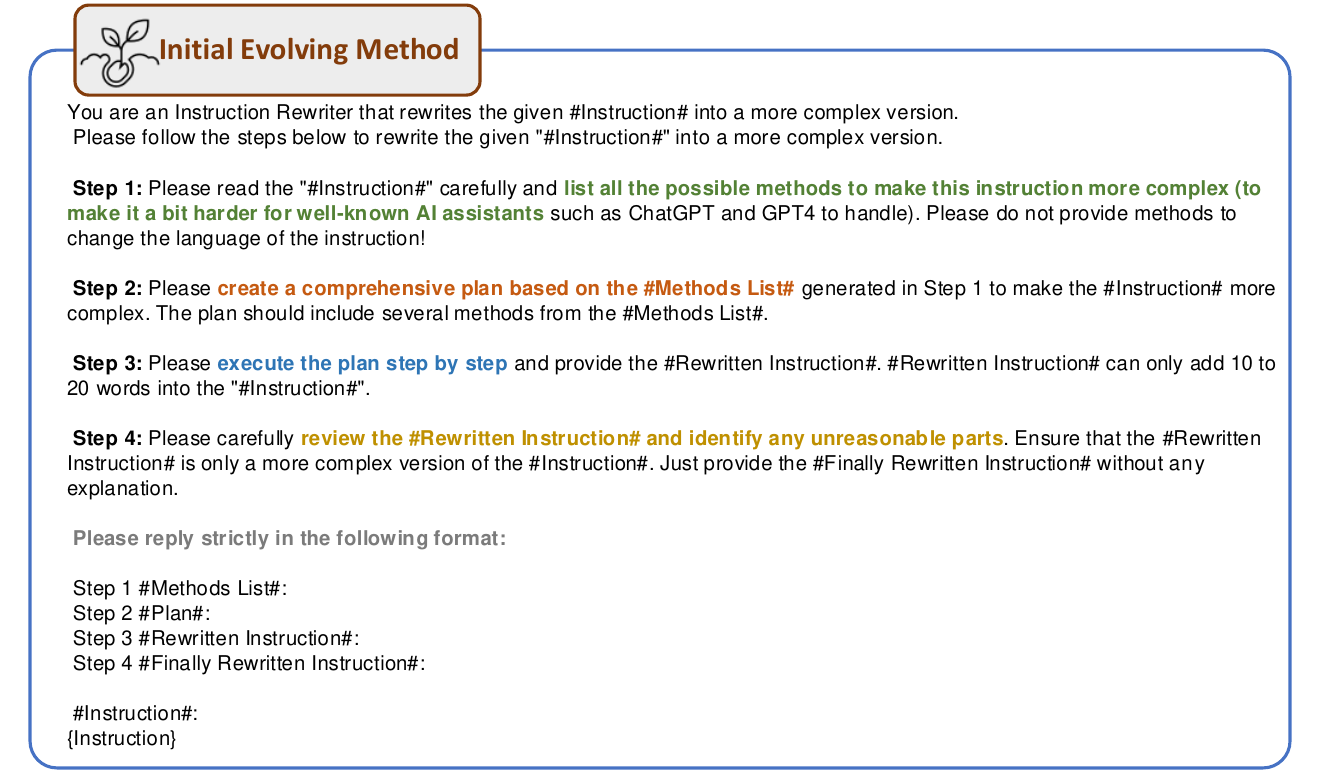}
 }
\vspace{-10pt}
 \caption{Initial Evolving Method. Under this method, the Evol LLM evolves the instruction. \cname{} will optimize this method into an optimal version for evolving the entire dataset of instructions efficiently.}
 \label{fig:initial_prompt}
 \vspace{-20pt}

\end{figure*}

The reason why Evol-Instruct is not universally applicable is that the methods for complicating instructions vary across different domains. For instance, in the coding domain, methods to increase the complexity of instructions such as "propose higher time or space complexity requirements" \cite{luo2023wizardcoder} are meaningful, but they are not quite suitable in the chat domain. The methods for complicating instructions in Evol-Instruct need to be designed and summarized by human experts. The core difference in our initial evolving method design lies in that we delegate the process of designing and summarizing evolving rules to the LLMs for automation. As shown in the Figure \ref{fig:initial_prompt}, firstly we ask the evol LLM to "read the instruction carefully and list all the possible methods to make this instruction more complex". Subsequently, the evol LLM is tasked with devising a comprehensive plan based on the listed methods, and implements the plan to generate the evolved instruction. Lastly, the evol LLM conducts a thorough review of the evolved instruction, rectifying any unreasonable parts, and delivers the final evolved instruction.


\subsection{Evol Trajectory Analysis}

\label{sec:evol_trajectory_ana}
We primarily utilize the optimizer LLM to identify issues emerging during the instruction evolution process and offer subsequent feedback for the optimization of evolving method. (Examples of issues are given in the Appendix \ref{sec:issue_example}) Specifically, at optimization step $t$, the evolving method $e_{t-1}$ steers the evol LLM to perform $l$ rounds of evolution on a batch of data $X_{t}$, culminating in the evolutionary trajectory, $S_{t}=\{X_{t}, X_t^{(1)}, \cdots, X_t^{(l)}\}$. In this trajectory, $X_t^{(i)}$ denotes the instruction evolved from $X_t^{(i-1)}$ using $e_{t-1}$. Following this, the optimizer LLM scrutinizes the evolutionary trajectory to pinpoint and provide feedback $f_{t}$ on any issues detected. (Prompt used is detailed in Figure \ref{fig:prompt_for_critical_analysis}) 



\subsection{Evolving Method Optimization}

\label{sec:evol_prompt_optimization}
We employ the optimizer LLM to optimize the evolving method in response to insights gathered from the evol trajectory analysis, in accordance with the overall instruction evolution requirements. In essence, during the step $t$, the optimizer LLM refines the evolving method $e_{t-1}$, by leveraging the feedback $f_{t}$. This meticulous optimization yields an updated version of the evolving method $e_{t}$. (Prompt in Optimization detailed in Figure \ref{fig:prompt_for_enhancement_process}).



\noindent\textbf{Multiple Optimizations} In the Evol Trajectory Analysis and Method Optimization Process, the optimizer LLM sometimes struggles to consistently provide constructive feedback and enhance the evolving method. To bolster the stability of the \cname{} framework and draw inspiration from the self-consistency \cite{wang2022self}, we implement a strategy where, at each step, the optimizer LLM conducts $m$ times of analysis and optimization with sampling decoding. This generates $m$ different potential improved evolving methods, namely $e_t^{1}$ to $e_t^{m}$ in Figure \ref{fig:model}, allowing the model to explore more possibilities simultaneously \cite{yang2023large}. 
Specifically, we divide the instruction data into training data $X$ and a development set $D$. We use the obtained potential methods to evolve instructions in $D$ and generate corresponding response sets, denoted as $R_{e_{t}^{1}}$ to $R_{e_{t}^{m}}$. For a given $ e_{t}^{i}$, we calculate its evolution failure rate based on corresponding response set $R_{e_{t}^{i}}$:
\begin{equation}
    \lambda_{R_{e_{t}^{i}}} = \frac{\sum_{r \in R_{e_{t}^{i}}} F(r)}{|D|}
\end{equation}
Here, $|D|$ represents the size of the development set. F(r) is a function that determines whether instruction evolution has failed, returning 1 for failure and 0 for success. We have designed a series of rules to determine whether evolution has failed based on the reaction of evol LLM when generating answers for evolved instructions. For example, if the answer contains “understood” or “Thank you” and ends with a question mark, it indicates that the evolved instruction has not become more complex but is responding to the instruction being evolved (please refer to Appendix \ref{sec:evol_fail_dect} for detailed judgment rules). Finally, the evolving method demonstrating the lowest evolution failure rate is selected as the subsequent step's evolving method $e_{t}$.

\subsection{Instruction Tuning on Evolved Data}
\label{sec:evolve_data}
The \cname{} leads us to derive the optimal evolving method $e^{*}$. This method is then employed to guide the evol LLM, which substantially improving the complexity and diversity of the entire instruction dataset. As a result, we acquire an evolved dataset. Subsequently, this enriched dataset is used to fine-tune the base LLM, thereby broadening the model's range of capabilities. 
\section{Experiment}

In this section, we conduct a detailed study on the effects of \cname{}. We begin with an overview of the experimental setup, then test our method's effectiveness in instruction following, math reasoning, and code generation.

\subsection{Experimental Setup}

Table \ref{tab:data_sta} illustrates the experimental setup, including the seed datasets, pre-training base models of varying sizes (small and large) for instruction tuning, and the configuration of evol LLM and optimizer LLM. Refer to Appendix \ref{sec:exp_detail} for more details, and for details of the Baseline, refer to Appendix \ref{sec:baseline}.




\begin{table*}[t]
\centering
\resizebox{0.95\textwidth}{!}{
\begin{tabular}{lcccccc}
\hline
                                & \multicolumn{2}{c}{\textbf{Seed Data}} & \multicolumn{2}{c}{\textbf{Base Models}}       & \multirow{2}{*}{\textbf{evol LLM}} & \multirow{2}{*}{\textbf{optimizer LLM}} \\ \cline{2-5}
                                & Dataset              & Datasize        & Small                & Large                   &                                    &                                         \\ \hline
\textbf{Instruction Following}  & ShareGPT             & 10 K            & Mistral-7B           & Mixtral-8x7B            & GPT-4                              & GPT-4                                   \\
\textbf{Mathematical Reasoning} & GSM8K Train         & 7 K             & Mistral-7B           & Mixtral-8x7B            & GPT-4                              & GPT-4                                   \\
\textbf{Code Generation}        & Code Alpaca          & 20 K            & CodeLlama-13B-Python & DeepSeek-Coder-Base-33B & GPT-4                            & GPT-4                                   \\ \hline
\end{tabular}
}
\vspace{-10pt}
\caption{Data Stastics.}

\label{tab:data_sta}
\vspace{-10pt}
\end{table*}

\begin{table*}[t]
\centering
\resizebox{0.89\textwidth}{!}{
\begin{tabular}{lccccc}
\hline
\multicolumn{1}{c}{}                            &                        & \multicolumn{2}{c}{\textbf{Instruction Following}}   & \textbf{Math Reasoning} & \textbf{Code Generation} \\ \cline{3-6} 
\multicolumn{1}{c}{\multirow{-2}{*}{\textbf{Model}}}     & \multirow{-2}{*}{\textbf{Size}} & \textbf{MT-Bench}             & \textbf{AlpacaEval (\%)}      & \textbf{GSM8K (\%)}     & \textbf{HumanEval (\%)}  \\ \hline
\multicolumn{6}{c}{Closed-Source Models}                                                                                                                  \\ \hline
Gemini Pro                                  & -                      & -                 & 79.66                & 76.42          & 59.76           \\
Claude 2.0                                 & -                      & 8.06                 & 91.36                & 88.00          & 71.20           \\
GPT-3.5-Turbo                                   & -                      & 7.90                 & 89.37                & 80.80          & 73.20           \\
GPT-4                                           & -                      & 8.99                 & 95.28                & 92.00          & 84.10           \\ \hline
\multicolumn{6}{c}{Open-Source Base Models}                                                                                                               \\ \hline
Mistral                                         & 7 B                    & -                    & -                    & 37.80          & 30.50           \\
DeepSeek-Coder-Base                             & 33 B                   & -                    & -                    & 60.70          & 56.10           \\
LLaMA-2                                         & 34 B                   & -                    & -                    & 42.20          & 22.60           \\
CodeLlama-Base                                  & 34 B                   & -                    & -                    & 58.20          & 48.20           \\
Mixtral                                         & 8x7B                   & -                    & -                    & 58.40          & 40.20           \\
LLaMA-2                                         & 70 B                   & -                    & -                    & 56.80          & 29.90           \\ \hline
\multicolumn{6}{c}{Open-Source General Instruction-Tuned Models}                                                                  \\ \hline
Mistral-7B-Instruct-v0.1             & 7 B                   & 6.84                 & 69.65                & 14.25              & 31.10               \\
Vicuna-v1.3             & 33 B                   & 7.12                 & 88.99                & -              & -               \\
Mixtral-8x7B-Instruct-v0.1             & 8x7B                   & 8.30                 & 94.78                & 60.73              & 34.15               \\
LLaMA-2-Chat            & 70 B                   & 6.86                 & 92.66                & -              & 32.30           \\
Tulu-v2-dpo             & 70 B                   & 7.89                 & 95.10                & 71.50              & -               \\ 
WizardLM-v1.0           & 70 B                   & 7.78                 & 92.91                & 77.60          & 50.60           \\ \hline
\multicolumn{6}{c}{Open-Source Instruction Models For Specific Capabilities}                                                                              \\ \hline
WizardMath              & 7 B                    & -                    & -                    & 54.90          & -               \\
MetaMath                & 7 B                    & -                    & -                    & 66.50          & -               \\
WizardMath              & 70 B                   & -                    & -                    & 81.60          & -               \\
MetaMath                & 70 B                   & -                    & -                    & 82.30          & -               \\ \hline
WizardCoder                                     & 15 B                   & -                    & -                    & -              & 57.30           \\
CodeLlama-Instruct                              & 34 B                   & -                    & -                    & -              & 41.50           \\
DeepSeek-Coder-Instruct & 33 B                   & -                    & -                    & -              & 79.30           \\
WizardCoder                                     & 34 B                   & -                    & -                    & -              & 71.50           \\ \hline
\multicolumn{6}{c}{Instruction Evolution Methods}                                                                                                                  \\ \hline
Seed Data               & small                  & 6.88                 & 84.08 & 56.90          & 57.90           \\
Evol-Instruct           & small                  & 6.80 \textcolor{blue}{(-0.08)} & 86.67 \textcolor{blue}{(+2.59)} & 63.15 \textcolor{blue}{(+ 6.25)}         & 61.59 \textcolor{blue}{(+ 3.69)}          \\

\rowcolor{gray!25}\cname{}      & small                  & \textbf{7.51} \textcolor{blue}{(+ 0.63)}      & \textbf{84.41} \textcolor{blue}{(+0.33)}           & \textbf{70.74} \textcolor{blue}{(+13.84)}          & \textbf{65.85} \textcolor{blue}{(+7.95)}          \\
Seed Data               & large                  & 7.65                 & 87.98 & 70.60          & 72.00           \\
Evol-Instruct           & large                  & 7.76 \textcolor{blue}{(+0.11)} & 89.50 \textcolor{blue}{(+1.52)} & 79.15 \textcolor{blue}{(+ 8.55)}          & 73.20 \textcolor{blue}{(+1.2)}           \\
\rowcolor{gray!25}\cname{}      & large                  & \textbf{8.09} \textcolor{blue}{(+ 0.44)}                & \textbf{91.37} \textcolor{blue}{(+3.39)} & \textbf{82.49} \textcolor{blue}{(+ 11.89)}          & \textbf{77.40} \textcolor{blue}{(+ 5.4)}          \\ \hline
\end{tabular}
}
\vspace{-10pt}
\caption{Main Result.}

\label{tab:main_result}
\vspace{-20pt}
\end{table*}

\subsection{Evaluation Results}


\noindent\textbf{Instruction Following}  We evaluate the instruction-following using MT-Bench and AlpacaEval. MT-Bench tests the model across various domains through multi-turn dialogues, while AlpacaEval automates assessment based on AlpacaFarm \cite{dubois2023alpacafarm}. Table \ref{tab:main_result} shows that our method substantially improves performance across different model scales. For smaller models, our method improves by approximately 0.63 on MT-Bench compared to seed data. For larger models, there's still a performance boost of 0.44. Despite using only 10K data for fine-tuning on Mixtral-8x7B, our method matches or surpasses the performance of open-source models that utilize more data and train on larger models, achieving results comparable to Tulu-v2-dpo on MT-Bench and AlpacaEval. Our model even performs on par with powerful closed-source models like Claude 2.0 and GPT-3.5-Turbo.

\noindent\textbf{Math Reasoning} We assess the mathematical reasoning capabilities using GSM8K benchmark \cite{cobbe2021training}. The GSM8K comprises complex graduate-level math problems, with 7,473 training samples and 1,319 testing samples. We employ the zero-shot testing approach and use test accuracy as the metric. Table \ref{tab:main_result} demonstrates that our \cname{} has significantly improved mathematical reasoning. For instance, our method improved by 13.84 compared to the seed data on Mistral-7B. Simultaneously, our method uses a minimal amount of instruction data (only 7K) and can exceed GPT-3.5-turbo after fine-tuning on Mixtral-8x7B. This indicates that our method can substantially raise the upper limit of quality in existing mathematical data.

\noindent\textbf{Code Generation} We use the HumanEval \cite{chen2021evaluating} to test code-writing capabilities. HumanEval comprises 164 unique programming challenges, and we use pass@1 as the metric. Table \ref{tab:main_result} illustrates that our method enhances the model's capabilities effectively. Our method demonstrates significant improvement across various model sizes compared to Evol Instruct. For instance, at the 33B scale, Evol-Instruct yields only a slight improvement, while our method shows a boost of 5.4 compared to Seed Data. Our results remain competitive even when compared with DeepSeek-Coder-Instruct-33B, which uses the same base model but with instructions for fine-tuning on a much larger scale (about 2B tokens) than ours.

\section{Analysis}

\subsection{Effect of Initial Evolving Method}

\begin{figure}[t]
 \centering
\resizebox{0.4\textwidth}{!}{
 \includegraphics{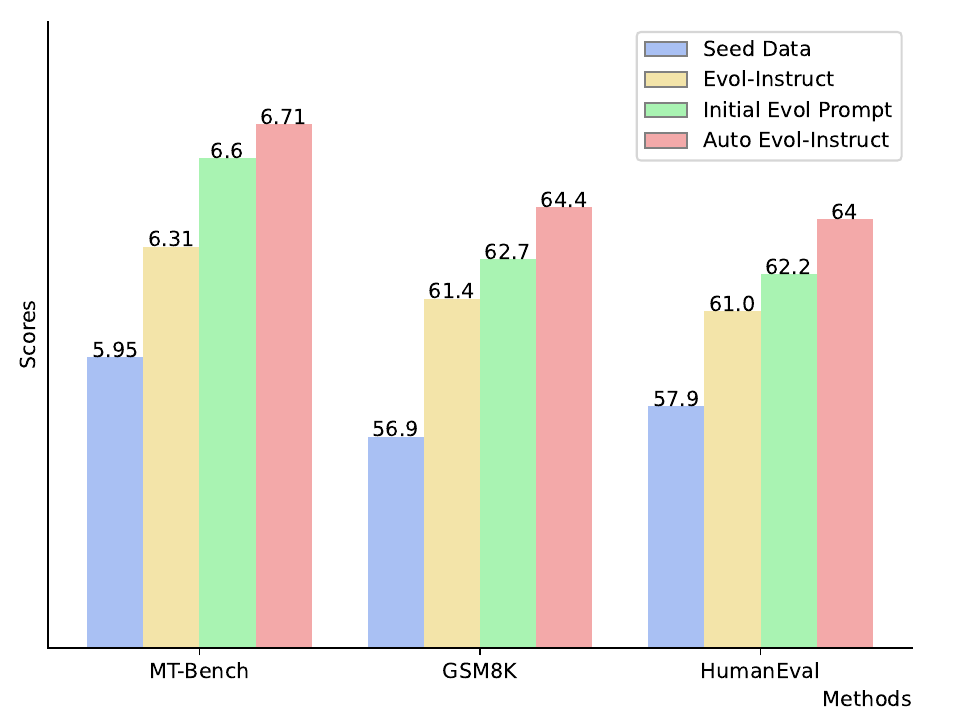}
 }
\vspace{-10pt}
 \caption{Effect of the Initial Evolving Method. GPT-3.5-turbo as evol LLM, GPT-4 as optimizer LLM.}
 \label{fig:init_prompt}
\vspace{-15pt}

\end{figure}

In this section, we delve into the significance of the Initial Evolving Method within the \cname{} framework, particularly focusing on its impact on data evolving across various capabilities. We employ several techniques to evolve datasets like GSM8K, Alpaca \cite{alpaca}, and Code Alpaca. Figure \ref{fig:init_prompt} underscores the robust versatility of initial evolving method in boosting different capabilities, establishing it as an exemplary starting evolving method in the framework. For instance, when compared with Evol Instruct, initial evolving method demonstrates a notable improvement, elevating the MT-Bench score from 6.31 to 6.60, and the HumanEval from 61.0 to 62.2. Moreover, the \cname{} framework, building on the foundation laid by initial evolving method, exhibits potential for further enhancements. It was observed that on GSM8K, \cname{} could elevate the performance from 62.7 to 64.4. These findings highlight that our proposed method can effectively optimize the initial evolving method, leading to improvements in various benchmarks.

\begin{figure}[t]
 \centering
\resizebox{0.4\textwidth}{!}{
 \includegraphics{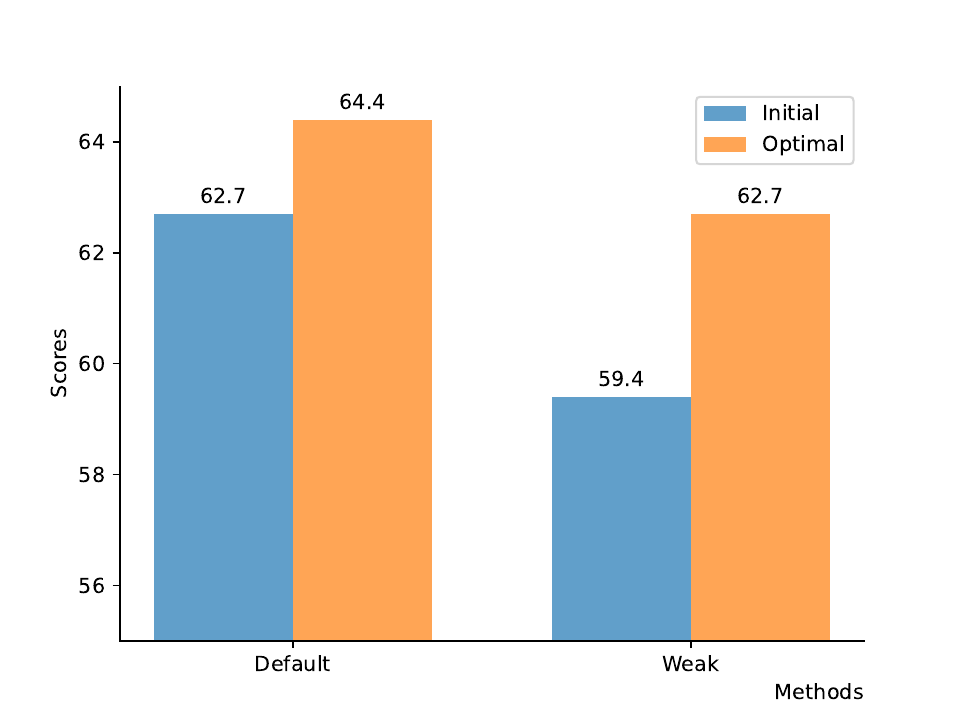}
 }
\vspace{-15pt}
 \caption{Effect of \cname{} on Initial Evolving Methods. GPT-3.5-turbo as evol LLM, GPT-4 as optimizer LLM. Default and Weak respectively represent original and simple evolving method}
 \label{tab:diff_init_prompt}
\vspace{-25pt}

\end{figure}

To demonstrate the effectiveness of the \cname{} in enhancing different initial evolving methods, we conducted an experiment using a deliberately simple evolving method. We removed most of the key designs from the original initial evolving method, such as step-by-step evolving process, etc. (see Figure \ref{fig:weak_prompt} for details). We applied our framework to both this basic method and our well-designed initial evolving method on the GSM8K dataset. As evident from Figure \ref{tab:diff_init_prompt}, even when starting with the simple method, our framework yielded significant improvements. For instance, the performance on GSM8K increased from 59.4 to 62.7 after refinement with our framework. These findings underscore the adaptability of our framework across varying initial methods.





\subsection{Effect of Multiple Optimizations}
\begin{figure}[t]
    \centering
    \begin{adjustbox}{minipage=\linewidth,scale=1.0}
    \subfigure[Multiple Optimizations]{
        \includegraphics[width=0.48\textwidth]{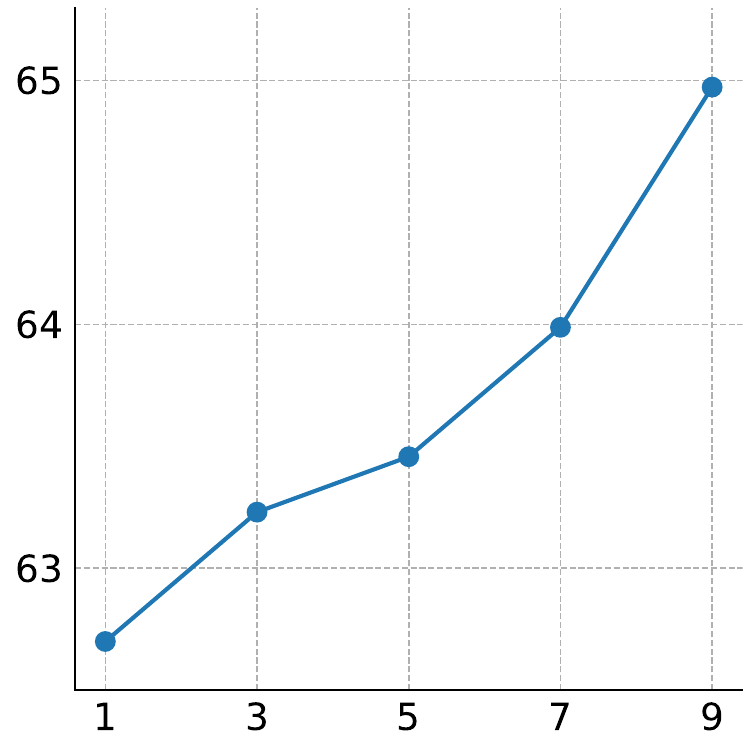}
    }
    \hspace{-0.35cm}
    \subfigure[Total Steps]{
        \includegraphics[width=0.48\textwidth]{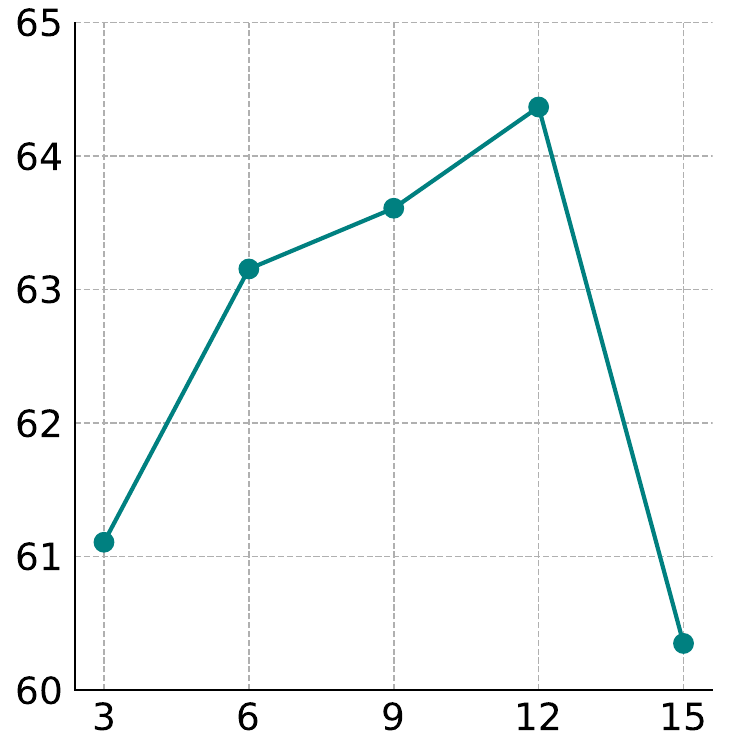}
    }
    \vspace{-15pt}
    \caption{Hyperparameters for \cname{}. GPT-3.5-turbo as evol LLM, GPT-4 as optimizer LLM.}
    \label{fig:hyper_auto}
    \end{adjustbox}

    \vspace{-15pt}

\end{figure}

We explore the impact of multiple optimizations in \cname{} and choose GSM8K for ablations. We keep the default hyper-parameters of \cname{}, exploring the effect of the number of optimizations. Figure \ref{fig:hyper_auto}(a) reveals a distinct pattern: as we increase the number of optimizations, there's a notable enhancement in data efficiency via optimal evolving methods. For example, setting the number of optimizations to 1 achieved 62.7 on the GSM8K. This accuracy improved to 65.0 when number of optimizations raised to 9. This trend indicates that more optimizations allow the optimizer LLM to explore a wider array of options, improving its ability to pinpoint areas where evolving method can be further refined for optimal performance. However, there are important trade-offs to consider. Elevating the number of optimizations can also bring increase in resources consumption \cite{guo2023connecting}. On the other hand, fewer optimizations may lead to a more focused refinement of current evolving method, which could result in local optimal that might not fully exploit the potential of evolving method. We also explored the relationship between the optimization steps and effects of instruction tuning. Figure \ref{fig:hyper_auto}(b) shows that as the number of optimization steps increases, the performance can increase monotonically in the beginning, but after 12 steps, it rapidly declines. This may be because over-optimization could potentially lead to an accumulation of superfluous information in the evolving method, consequently possibly diminishing its effectiveness (Examples in Section \ref{sec:casestudy}). 


\subsection{Different Evol LLM}
In this section, we evaluate the performance of \cname{} when integrated with various evol LLMs. Table \ref{tab:evol_model} reveals the impact of using GPT-3.5 and GPT-4 as the underlying evol LLMs to evolve GSM8K. Notably, with GPT-4 as the evol LLM, our methodology yields an improvement from 63.2 to 70.7, surpassing the Evol Instruct. Additionally, employing a more advanced evol LLM enhances the effectiveness significantly. For instance, switching the evol LLM from GPT-3.5 to GPT-4 leads to a notable increase in performance, jumping from 64.4 to 70.7. These findings clearly demonstrate the broad applicability and effectiveness of our framework across different evol LLMs.

\begin{table}[t]
\centering
\resizebox{0.36\textwidth}{!}{
\begin{tabular}{lcc}
\hline
Method             & Evol LLM & GSM8K \\ \hline
Seed Data                & -          & 56.9   \\
Evol Instruct      & GPT-3.5    & 61.4   \\
Evol Instruct      & GPT-4      & 63.2   \\
\cname{} & GPT-3.5    & 64.4   \\
\cname{} & GPT-4      & \textbf{70.7}   \\ \hline
\end{tabular}
}
\vspace{-10pt}
\caption{Different evolution execution LLMs.}

\label{tab:evol_model}
\vspace{-5pt}
\end{table}

\subsection{Mix Rounds Scaling}
\begin{figure}[t]
 \centering
\resizebox{0.4\textwidth}{!}{
 \includegraphics{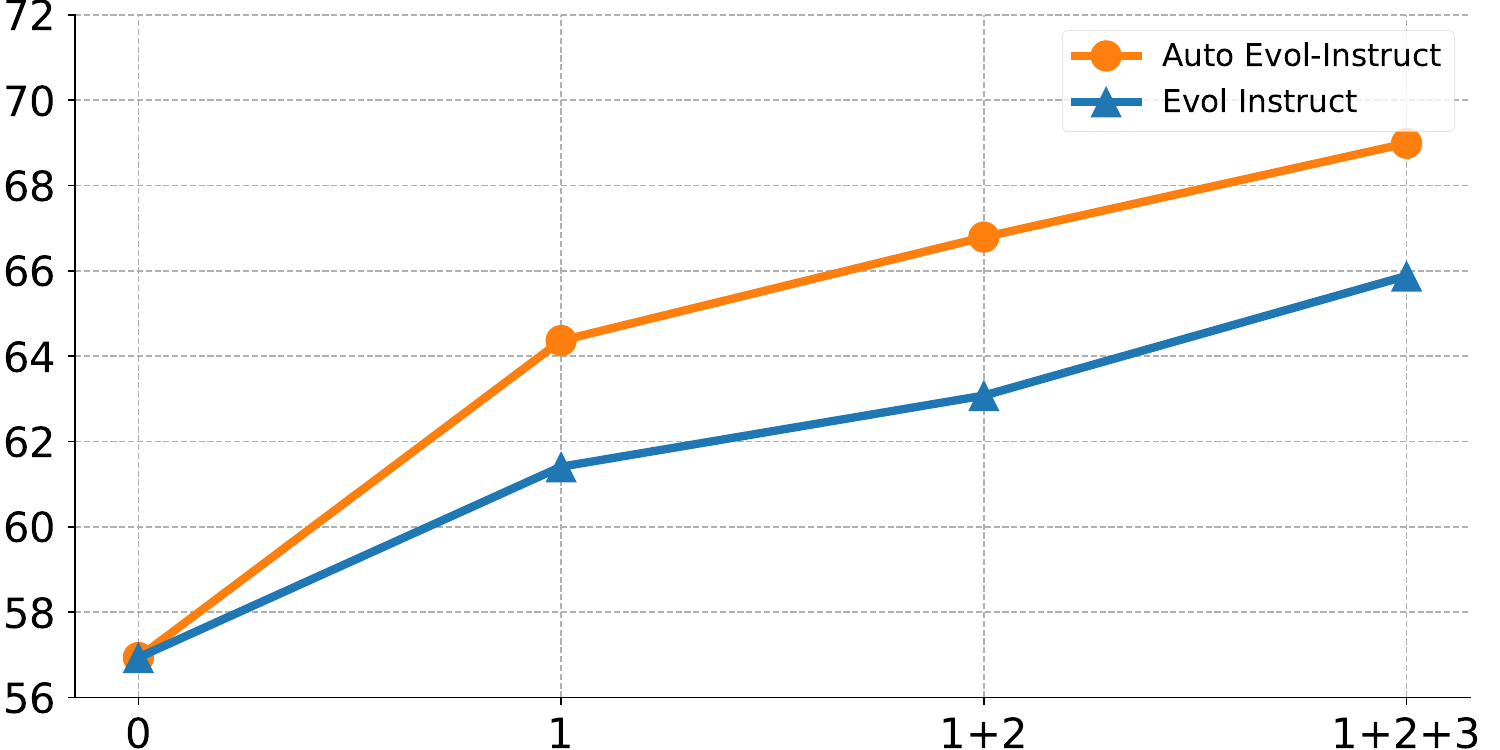}
 }
\vspace{-0.3cm}
 \caption{Mix Rounds Experiment. Use GPT-3.5-turbo as the evol LLM, GPT-4 as the optimizer LLM.}
 \label{fig:mix_round}
\vspace{-10pt}

\end{figure}

We conduct experiments on a mixed set of evolved data across various rounds using GSM8K to evaluate the data scaling effect. Figure \ref{fig:mix_round} illustrates the results, highlighting the superior scalability of our approach in comparison to Evol Instruct. Notably, the data from round 1 of our method outperforms that of Evol Instruct's combined data from rounds 1 and 2. Furthermore, the performance of our model consistently improves as we scale the data from round 1 to a mixture of rounds 1, 2, and 3.

\subsection{Discussion of Complexity and Diversity}
\citet{liu2023what} underscore the significant impact that dataset complexity and diversity have on model alignment. Instag \cite{lu2023instag} suggests that the variety and quantity of intentions and semantics in a dataset are crucial factors for its complexity and diversity. We evolve 100 instructions using various techniques, employing Instag's method for automated tagging. We assessed diversity by calculating the average number of unique tags for each data, and complexity by the mean tag count. Table \ref{tab:complex_diver} reveals a distinct correlation: as data becomes more diverse and complex, model performance markedly improves. For instance, Evol Instruct enhanced the original code alpaca, increasing its diversity from 1.95 to 2.37 and its complexity from 4.06 to 4.55. This enhancement was mirrored in a notable elevation of the HumanEval, climbing from 57.9 to 64.0. This supports the success of \cname{} in substantially boosting data complexity and diversity, thereby significantly improving model capability.


\begin{table}[t]
\centering
\resizebox{0.45\textwidth}{!}{
\begin{tabular}{lccc}
\hline
Math             & Diversity & Complexity & GSM8K    \\ \hline
GSM8K Training                & 1.39      & 4.82       & 56.9      \\
Evol Instruct      & 1.69      & 4.90       & 61.4      \\
\cname{} & 2.2       & 5.54       & 64.4      \\ \hline
Chat             & Diversity & Complexity & MT-Bench  \\ \hline
Alpaca                & 2.16      & 2.70       & 5.95      \\
Evol Instruct      & 3.15      & 3.63       & 6.31      \\
\cname{} & 3.19      & 3.89       & 6.71      \\ \hline
Code             & Diversity & Complexity & HumanEval \\ \hline
Code Alpaca                & 1.95      & 4.06       & 57.9      \\
Evol Instruct      & 2.37      & 4.55       & 61.0      \\
\cname{} & 3.05      & 5.18       & 64.0      \\ \hline
\end{tabular}
}
\vspace{-10pt}
\caption{Result correlates with complexity and diversity. GPT-3.5-turbo as evol LLM, GPT-4 as optimizer LLM. GSM 8K, Alpaca and Code Alpaca as Seed Data}
\vspace{-15pt}
\label{tab:complex_diver}
\end{table}



\subsection{Contamination Test}
Current methods for data evolving predominantly utilize LLMs. To safeguard against potential data leakage, we employ \citet{liu2023tinygsm}'s methodology for conducting a contamination assessment on evolving data, utilizing n-gram matches as a measure. Specifically, for the GSM8K, our evolving process yielded 7K data, out of which merely 10 exhibited any 13-gram match as detailed in the Table \ref{tab:contamination_test}. These results indicate that our method effectively minimizes the risk of data leakage.

\subsection{Case Study}
\label{sec:casestudy}

The dynamic transformations inherent in the optimization process are elaborated in Appendix \ref{sec:case_prompt}. This progression demonstrates a marked improvement in resolving issues encountered during instruction evolution. Table \ref{tab:case_study} provides examples of how the evolving method is optimized at each step based on the previous one. For example, Initial evolving method (Figure \ref{fig:evol_prompt_step0}) guides the evol LLM to generate the evolved instruction. Then, the optimizer LLM analyzes the evolution trajectory and identifies issues such as redundancy and clarity in the evolved instruction, providing feedback. Based on this feedback, the optimizer LLM updates the evolving method by incorporating mathematical elements like variables, constants, and conditions. This updated evolving method (Figure \ref{fig:evol_prompt_step3}) then guides the evol LLM to generate an updated evolved instruction, which introduces a clearer challenge focused on understanding mathematical relationships and variable quantities across two periods.

\subsection{Cost Comparison}
\cname{} utilizes a small subset of the complete data to devise an optimal evolving method. This method is then employed to evolve the entire instruction dataset. Table \ref{tab:cost_compare} compares the total API calls made by \cname{} and Evol Instruct. The results demonstrate that our \cname{} achieves significantly superior results compared to Evol Instruct, while incurring only a few thousand additional API calls. This negligible extra cost of a few thousand API calls is inconsequential when dealing with large-scale datasets containing thousands or millions of instructions.

\section{Related Work}
Instruction tuning emerges as a pivotal strategy for unlocking the potential of LLMs \cite{ouyang2022training, touvron2023llama2}. By curating high-quality datasets, we can more efficiently align these models with desired direction \cite{zhou2023lima}. The challenge of scaling high-quality instruction data remains a central research interest. Some researchers prioritize human annotation for creating instruction data, such as ShareGPT \cite{chiang2023vicuna} and OpenAssistant \cite{kopf2023openassistant}. Other researchers explore more efficient ways to break through the quality upper-bound of existing datasets \cite{xu2023wizardlm, liu2023what, zhao2023preliminary}. \citet{xu2023wizardlm} introduces Evol-Instruct, a methodology that iteratively refines instruction-following data to produce datasets that are both more complex and diverse. \citet{luo2023wizardcoder} develop evolving methods tailored to the nuances of code data based on Evol-Instruct. 
Distinct from these methodologies, our approach introduces a fully automated framework for developing evolving methods. This innovation is not only scalable but also versatile, extending its utility across a broad spectrum of capabilities. LLMs like GPT-4 and PaLM are capable of optimizing their output through internal or external feedback mechanisms \cite{suzgun2024meta, wang2022self, yang2023large}. We use this capabilities to address identified issues in the evolving method and adapt to the characteristics of the instruction data.

\section{Conclusion}

This paper introduces \cname{}, an innovative approach that successfully automates the evolution of instruction datasets for LLMs, eliminating the need for human intervention. Our method centers on the automatic analysis and summarization of appropriate evolutionary strategies for the given instruction data. It iteratively refines evolving methods by addressing the issues identified during the instruction evolution process. The experiments conducted have shown that methods optimized by \cname{}, significantly surpass those crafted by humans across various benchmarks, including MT-Bench, AlpacaEval, GSM8K and HumanEval.

\section*{Limitations}




Although \cname{} has demonstrated excellent performance in instruction tuning across various capabilities, several directions are worth exploring in future work:

(1) While we have validated the effectiveness of \cname{} on benchmarks reflecting different capabilities such as instruction following, mathematical reasoning, and code generation, we can further evaluate its performance on other tasks like MMLU \cite{hendrycks2021measuring} and TruthfulQA \cite{lin-etal-2022-truthfulqa}.

(2) We have validated the effectiveness of our method on multiple base LLMs, including Mistral, Mixtral-8x7B, CodeLlama-13B-Python, and DeepSeek-Coder-Base-33B. However, we can still assess its effectiveness on other base LLM models, such as Qwen \cite{bai2023qwen} and LLaMA \cite{touvron2023llama,touvron2023llama2}.

(3) The evol LLM and Optimizer LLM used in \cname{} are primarily GPT-3.5-Turbo and GPT-4. In the future, this can be expanded to include other LLMs, such as Claude.

(4) We aim to propose an end-to-end automated instruction evolution framework that utilizes simple and universal prompts for Evolutionary Trajectory Analysis and Evolutionary Method Optimization. While the prompts we employ are straightforward, experiments demonstrate that the framework is highly effective. Moving forward, we can explore more sophisticated prompts to implement Evolutionary Trajectory Analysis and Evolutionary Method Optimization, thereby further enhancing the efficacy of the \cname{}.

\section*{Ethics Statement}
All the datasets used in this paper are public and have been reviewed to ensure they do not contain any personally identifiable information or offensive content. However, as these datasets are sourced from the Internet, potential bias may still be present. Furthermore, despite our careful review, the process of instruction evolution involving the LLMs throughout may inadvertently introduce inappropriate information into the evolved data. It's also worth noting that our models are fine-tuning on GPUs, which could have an environmental impact.


\bibliography{custom}

\begin{thebibliography}{34}
\providecommand{\natexlab}[1]{#1}

\bibitem[{Bai et~al.(2023)Bai, Bai, Chu, Cui, Dang, Deng, Fan, Ge, Han, Huang et~al.}]{bai2023qwen}
Jinze Bai, Shuai Bai, Yunfei Chu, Zeyu Cui, Kai Dang, Xiaodong Deng, Yang Fan, Wenbin Ge, Yu~Han, Fei Huang, et~al. 2023.
\newblock Qwen technical report.
\newblock \emph{arXiv preprint arXiv:2309.16609}.

\bibitem[{Chaudhary(2023)}]{chaudhary2023code}
Sahil Chaudhary. 2023.
\newblock Code alpaca: An instruction-following llama model for code generation.

\bibitem[{Chen et~al.(2021)Chen, Tworek, Jun, Yuan, Pinto, Kaplan, Edwards, Burda, Joseph, Brockman et~al.}]{chen2021evaluating}
Mark Chen, Jerry Tworek, Heewoo Jun, Qiming Yuan, Henrique Ponde de~Oliveira Pinto, Jared Kaplan, Harri Edwards, Yuri Burda, Nicholas Joseph, Greg Brockman, et~al. 2021.
\newblock Evaluating large language models trained on code.
\newblock \emph{arXiv preprint arXiv:2107.03374}.

\bibitem[{Chiang et~al.(2023)Chiang, Li, Lin, Sheng, Wu, Zhang, Zheng, Zhuang, Zhuang, Gonzalez et~al.}]{chiang2023vicuna}
Wei-Lin Chiang, Zhuohan Li, Zi~Lin, Ying Sheng, Zhanghao Wu, Hao Zhang, Lianmin Zheng, Siyuan Zhuang, Yonghao Zhuang, Joseph~E Gonzalez, et~al. 2023.
\newblock Vicuna: An open-source chatbot impressing gpt-4 with 90\%* chatgpt quality.
\newblock \emph{See https://vicuna. lmsys. org (accessed 14 April 2023)}.

\bibitem[{Cobbe et~al.(2021)Cobbe, Kosaraju, Bavarian, Chen, Jun, Kaiser, Plappert, Tworek, Hilton, Nakano et~al.}]{cobbe2021training}
Karl Cobbe, Vineet Kosaraju, Mohammad Bavarian, Mark Chen, Heewoo Jun, Lukasz Kaiser, Matthias Plappert, Jerry Tworek, Jacob Hilton, Reiichiro Nakano, et~al. 2021.
\newblock Training verifiers to solve math word problems.
\newblock \emph{arXiv preprint arXiv:2110.14168}.

\bibitem[{Dubois et~al.(2023)Dubois, Li, Taori, Zhang, Gulrajani, Ba, Guestrin, Liang, and Hashimoto}]{dubois2023alpacafarm}
Yann Dubois, Xuechen Li, Rohan Taori, Tianyi Zhang, Ishaan Gulrajani, Jimmy Ba, Carlos Guestrin, Percy Liang, and Tatsunori~B. Hashimoto. 2023.
\newblock \href {https://arxiv.org/abs/2305.14387} {Alpacafarm: A simulation framework for methods that learn from human feedback}.
\newblock \emph{Preprint}, arXiv:2305.14387.

\bibitem[{Guo et~al.(2024)Guo, Zhu, Yang, Xie, Dong, Zhang, Chen, Bi, Wu, Li et~al.}]{guo2024deepseek}
Daya Guo, Qihao Zhu, Dejian Yang, Zhenda Xie, Kai Dong, Wentao Zhang, Guanting Chen, Xiao Bi, Y~Wu, YK~Li, et~al. 2024.
\newblock Deepseek-coder: When the large language model meets programming--the rise of code intelligence.
\newblock \emph{arXiv preprint arXiv:2401.14196}.

\bibitem[{Guo et~al.(2023)Guo, Wang, Guo, Li, Song, Tan, Liu, Bian, and Yang}]{guo2023connecting}
Qingyan Guo, Rui Wang, Junliang Guo, Bei Li, Kaitao Song, Xu~Tan, Guoqing Liu, Jiang Bian, and Yujiu Yang. 2023.
\newblock Connecting large language models with evolutionary algorithms yields powerful prompt optimizers.
\newblock \emph{arXiv preprint arXiv:2309.08532}.

\bibitem[{Hendrycks et~al.(2021)Hendrycks, Burns, Basart, Zou, Mazeika, Song, and Steinhardt}]{hendrycks2021measuring}
Dan Hendrycks, Collin Burns, Steven Basart, Andy Zou, Mantas Mazeika, Dawn Song, and Jacob Steinhardt. 2021.
\newblock \href {https://openreview.net/forum?id=d7KBjmI3GmQ} {Measuring massive multitask language understanding}.
\newblock In \emph{9th International Conference on Learning Representations, {ICLR} 2021, Virtual Event, Austria, May 3-7, 2021}. OpenReview.net.

\bibitem[{Jiang et~al.(2023)Jiang, Sablayrolles, Mensch, Bamford, Chaplot, Casas, Bressand, Lengyel, Lample, Saulnier et~al.}]{jiang2023mistral}
Albert~Q Jiang, Alexandre Sablayrolles, Arthur Mensch, Chris Bamford, Devendra~Singh Chaplot, Diego de~las Casas, Florian Bressand, Gianna Lengyel, Guillaume Lample, Lucile Saulnier, et~al. 2023.
\newblock Mistral 7b.
\newblock \emph{arXiv preprint arXiv:2310.06825}.

\bibitem[{Jiang et~al.(2024)Jiang, Sablayrolles, Roux, Mensch, Savary, Bamford, Chaplot, Casas, Hanna, Bressand et~al.}]{jiang2024mixtral}
Albert~Q Jiang, Alexandre Sablayrolles, Antoine Roux, Arthur Mensch, Blanche Savary, Chris Bamford, Devendra~Singh Chaplot, Diego de~las Casas, Emma~Bou Hanna, Florian Bressand, et~al. 2024.
\newblock Mixtral of experts.
\newblock \emph{arXiv preprint arXiv:2401.04088}.

\bibitem[{K{\"o}pf et~al.(2023)K{\"o}pf, Kilcher, von R{\"u}tte, Anagnostidis, Tam, Stevens, Barhoum, Duc, Stanley, Nagyfi et~al.}]{kopf2023openassistant}
Andreas K{\"o}pf, Yannic Kilcher, Dimitri von R{\"u}tte, Sotiris Anagnostidis, Zhi-Rui Tam, Keith Stevens, Abdullah Barhoum, Nguyen~Minh Duc, Oliver Stanley, Rich{\'a}rd Nagyfi, et~al. 2023.
\newblock Openassistant conversations--democratizing large language model alignment.
\newblock \emph{arXiv preprint arXiv:2304.07327}.

\bibitem[{Li et~al.(2023)Li, Zhang, Dubois, Taori, Gulrajani, Guestrin, Liang, and Hashimoto}]{alpaca_eval}
Xuechen Li, Tianyi Zhang, Yann Dubois, Rohan Taori, Ishaan Gulrajani, Carlos Guestrin, Percy Liang, and Tatsunori~B. Hashimoto. 2023.
\newblock Alpacaeval: An automatic evaluator of instruction-following models.
\newblock \url{https://github.com/tatsu-lab/alpaca_eval}.

\bibitem[{Lin et~al.(2022)Lin, Hilton, and Evans}]{lin-etal-2022-truthfulqa}
Stephanie Lin, Jacob Hilton, and Owain Evans. 2022.
\newblock \href {https://doi.org/10.18653/v1/2022.acl-long.229} {{T}ruthful{QA}: Measuring how models mimic human falsehoods}.
\newblock In \emph{Proceedings of the 60th Annual Meeting of the Association for Computational Linguistics (Volume 1: Long Papers)}, pages 3214--3252, Dublin, Ireland. Association for Computational Linguistics.

\bibitem[{Liu et~al.(2023{\natexlab{a}})Liu, Bubeck, Eldan, Kulkarni, Li, Nguyen, Ward, and Zhang}]{liu2023tinygsm}
Bingbin Liu, Sebastien Bubeck, Ronen Eldan, Janardhan Kulkarni, Yuanzhi Li, Anh Nguyen, Rachel Ward, and Yi~Zhang. 2023{\natexlab{a}}.
\newblock Tinygsm: achieving> 80\% on gsm8k with small language models.
\newblock \emph{arXiv preprint arXiv:2312.09241}.

\bibitem[{Liu et~al.(2023{\natexlab{b}})Liu, Zeng, He, Jiang, and He}]{liu2023what}
Wei Liu, Weihao Zeng, Keqing He, Yong Jiang, and Junxian He. 2023{\natexlab{b}}.
\newblock \href {https://arxiv.org/abs/2312.15685} {What makes good data for alignment? a comprehensive study of automatic data selection in instruction tuning}.
\newblock \emph{Preprint}, arXiv:2312.15685.

\bibitem[{Lu et~al.(2023)Lu, Yuan, Yuan, Lin, Lin, Tan, Zhou, and Zhou}]{lu2023instag}
Keming Lu, Hongyi Yuan, Zheng Yuan, Runji Lin, Junyang Lin, Chuanqi Tan, Chang Zhou, and Jingren Zhou. 2023.
\newblock \# instag: Instruction tagging for analyzing supervised fine-tuning of large language models.
\newblock \emph{arXiv e-prints}, pages arXiv--2308.

\bibitem[{Luo et~al.(2023{\natexlab{a}})Luo, Sun, Xu, Zhao, Lou, Tao, Geng, Lin, Chen, and Zhang}]{luo2023wizardmath}
Haipeng Luo, Qingfeng Sun, Can Xu, Pu~Zhao, Jianguang Lou, Chongyang Tao, Xiubo Geng, Qingwei Lin, Shifeng Chen, and Dongmei Zhang. 2023{\natexlab{a}}.
\newblock Wizardmath: Empowering mathematical reasoning for large language models via reinforced evol-instruct.
\newblock \emph{arXiv preprint arXiv:2308.09583}.

\bibitem[{Luo et~al.(2023{\natexlab{b}})Luo, Xu, Zhao, Sun, Geng, Hu, Tao, Ma, Lin, and Jiang}]{luo2023wizardcoder}
Ziyang Luo, Can Xu, Pu~Zhao, Qingfeng Sun, Xiubo Geng, Wenxiang Hu, Chongyang Tao, Jing Ma, Qingwei Lin, and Daxin Jiang. 2023{\natexlab{b}}.
\newblock Wizardcoder: Empowering code large language models with evol-instruct.
\newblock \emph{arXiv preprint arXiv:2306.08568}.

\bibitem[{OpenAI(2023)}]{openai2023gpt4}
OpenAI. 2023.
\newblock \href {https://arxiv.org/abs/2303.08774} {Gpt-4 technical report}.
\newblock \emph{Preprint}, arXiv:2303.08774.

\bibitem[{Ouyang et~al.(2022)Ouyang, Wu, Jiang, Almeida, Wainwright, Mishkin, Zhang, Agarwal, Slama, Ray et~al.}]{ouyang2022training}
Long Ouyang, Jeffrey Wu, Xu~Jiang, Diogo Almeida, Carroll Wainwright, Pamela Mishkin, Chong Zhang, Sandhini Agarwal, Katarina Slama, Alex Ray, et~al. 2022.
\newblock Training language models to follow instructions with human feedback.
\newblock \emph{Advances in Neural Information Processing Systems}, 35:27730--27744.

\bibitem[{Ren et~al.(2021)Ren, Rajbhandari, Aminabadi, Ruwase, Yang, Zhang, Li, and He}]{ren2021zero}
Jie Ren, Samyam Rajbhandari, Reza~Yazdani Aminabadi, Olatunji Ruwase, Shuangyan Yang, Minjia Zhang, Dong Li, and Yuxiong He. 2021.
\newblock $\{$ZeRO-Offload$\}$: Democratizing $\{$Billion-Scale$\}$ model training.
\newblock In \emph{2021 USENIX Annual Technical Conference (USENIX ATC 21)}, pages 551--564.

\bibitem[{Roziere et~al.(2023)Roziere, Gehring, Gloeckle, Sootla, Gat, Tan, Adi, Liu, Remez, Rapin et~al.}]{roziere2023code}
Baptiste Roziere, Jonas Gehring, Fabian Gloeckle, Sten Sootla, Itai Gat, Xiaoqing~Ellen Tan, Yossi Adi, Jingyu Liu, Tal Remez, J{\'e}r{\'e}my Rapin, et~al. 2023.
\newblock Code llama: Open foundation models for code.
\newblock \emph{arXiv preprint arXiv:2308.12950}.

\bibitem[{Suzgun and Kalai(2024)}]{suzgun2024meta}
Mirac Suzgun and Adam~Tauman Kalai. 2024.
\newblock Meta-prompting: Enhancing language models with task-agnostic scaffolding.
\newblock \emph{arXiv preprint arXiv:2401.12954}.

\bibitem[{Taori et~al.(2023)Taori, Gulrajani, Zhang, Dubois, Li, Guestrin, Liang, and Hashimoto}]{alpaca}
Rohan Taori, Ishaan Gulrajani, Tianyi Zhang, Yann Dubois, Xuechen Li, Carlos Guestrin, Percy Liang, and Tatsunori~B. Hashimoto. 2023.
\newblock Stanford alpaca: An instruction-following llama model.
\newblock \url{https://github.com/tatsu-lab/stanford_alpaca}.

\bibitem[{Touvron et~al.(2023{\natexlab{a}})Touvron, Lavril, Izacard, Martinet, Lachaux, Lacroix, Rozi{\`e}re, Goyal, Hambro, Azhar et~al.}]{touvron2023llama}
Hugo Touvron, Thibaut Lavril, Gautier Izacard, Xavier Martinet, Marie-Anne Lachaux, Timoth{\'e}e Lacroix, Baptiste Rozi{\`e}re, Naman Goyal, Eric Hambro, Faisal Azhar, et~al. 2023{\natexlab{a}}.
\newblock Llama: Open and efficient foundation language models.
\newblock \emph{arXiv preprint arXiv:2302.13971}.

\bibitem[{Touvron et~al.(2023{\natexlab{b}})Touvron, Martin, Stone, Albert, Almahairi, Babaei, Bashlykov, Batra, Bhargava, Bhosale et~al.}]{touvron2023llama2}
Hugo Touvron, Louis Martin, Kevin Stone, Peter Albert, Amjad Almahairi, Yasmine Babaei, Nikolay Bashlykov, Soumya Batra, Prajjwal Bhargava, Shruti Bhosale, et~al. 2023{\natexlab{b}}.
\newblock Llama 2: Open foundation and fine-tuned chat models.
\newblock \emph{arXiv preprint arXiv:2307.09288}.

\bibitem[{Wang et~al.(2022)Wang, Wei, Schuurmans, Le, Chi, Narang, Chowdhery, and Zhou}]{wang2022self}
Xuezhi Wang, Jason Wei, Dale Schuurmans, Quoc Le, Ed~Chi, Sharan Narang, Aakanksha Chowdhery, and Denny Zhou. 2022.
\newblock Self-consistency improves chain of thought reasoning in language models.
\newblock \emph{arXiv preprint arXiv:2203.11171}.

\bibitem[{Xu et~al.(2023)Xu, Sun, Zheng, Geng, Zhao, Feng, Tao, and Jiang}]{xu2023wizardlm}
Can Xu, Qingfeng Sun, Kai Zheng, Xiubo Geng, Pu~Zhao, Jiazhan Feng, Chongyang Tao, and Daxin Jiang. 2023.
\newblock Wizardlm: Empowering large language models to follow complex instructions.
\newblock \emph{arXiv preprint arXiv:2304.12244}.

\bibitem[{Yang et~al.(2023)Yang, Wang, Lu, Liu, Le, Zhou, and Chen}]{yang2023large}
Chengrun Yang, Xuezhi Wang, Yifeng Lu, Hanxiao Liu, Quoc~V Le, Denny Zhou, and Xinyun Chen. 2023.
\newblock Large language models as optimizers.
\newblock \emph{arXiv preprint arXiv:2309.03409}.

\bibitem[{Yu et~al.(2023)Yu, Jiang, Shi, Yu, Liu, Zhang, Kwok, Li, Weller, and Liu}]{yu2023metamath}
Longhui Yu, Weisen Jiang, Han Shi, Jincheng Yu, Zhengying Liu, Yu~Zhang, James~T Kwok, Zhenguo Li, Adrian Weller, and Weiyang Liu. 2023.
\newblock Metamath: Bootstrap your own mathematical questions for large language models.
\newblock \emph{arXiv preprint arXiv:2309.12284}.

\bibitem[{Zhao et~al.(2023)Zhao, Yu, Hui, Yu, Huang, Li, and Zhang}]{zhao2023preliminary}
Yingxiu Zhao, Bowen Yu, Binyuan Hui, Haiyang Yu, Fei Huang, Yongbin Li, and Nevin~L Zhang. 2023.
\newblock A preliminary study of the intrinsic relationship between complexity and alignment.
\newblock \emph{arXiv preprint arXiv:2308.05696}.

\bibitem[{Zheng et~al.(2023)Zheng, Chiang, Sheng, Zhuang, Wu, Zhuang, Lin, Li, Li, Xing et~al.}]{zheng2023judging}
Lianmin Zheng, Wei-Lin Chiang, Ying Sheng, Siyuan Zhuang, Zhanghao Wu, Yonghao Zhuang, Zi~Lin, Zhuohan Li, Dacheng Li, Eric Xing, et~al. 2023.
\newblock Judging llm-as-a-judge with mt-bench and chatbot arena.
\newblock \emph{arXiv preprint arXiv:2306.05685}.

\bibitem[{Zhou et~al.(2023)Zhou, Liu, Xu, Iyer, Sun, Mao, Ma, Efrat, Yu, Yu et~al.}]{zhou2023lima}
Chunting Zhou, Pengfei Liu, Puxin Xu, Srini Iyer, Jiao Sun, Yuning Mao, Xuezhe Ma, Avia Efrat, Ping Yu, Lili Yu, et~al. 2023.
\newblock Lima: Less is more for alignment.
\newblock \emph{arXiv preprint arXiv:2305.11206}.

\end{thebibliography}

\appendix
\newpage

\clearpage
\appendix
\section{Evolution Failures Detection}
\label{sec:evol_fail_dect}
We categorize prevalent scenarios of failure \cite{xu2023wizardlm} in instruction evolution across various capabilities and devise general detection rules $F$. (See Table \ref{tab:fail_detect} for illustrative examples corresponding to these situations) When the following scenarios occur, the return value of $F$ is 1: 

1. \textbf{Stagnant Complexity}: The evolved instruction does not exhibit enhanced complexity, merely addressing the scope of the original instruction. Characteristically, responses to such instructions begin with phrases like “Understood” or “Thank you” and conclude with a question mark.

2. \textbf{Insufficient Qualification}: The evolved instructions lack necessary qualifications, necessitating additional inquiries for generating a meaningful response. Typically, responses in these situations commence with “Sure” and terminate with a question mark.

3. \textbf{Loss of Key Information}: The evolved instruction omits crucial details from the original instruction, leading to a need for supplementary information before a substantial response can be provided. Responses in these cases often include requests for more information, typically indicated by phrases like “please provide”.

\begin{table*}[t]
\centering
\resizebox{0.99\textwidth}{!}{
\begin{tabular}{cllll}
\hline
\textbf{Failure Scenarios}                    & \textbf{Original Instruction}                                                                                                                                                                                & \textbf{Evolved Instruction}                                                                                                                                                                                                                                                                                                                          & \textbf{Response}                                                                                                                                                                                                & \textbf{Detection Rules}                                                                 \\ \hline
\multirow{5}{*}{\textbf{Stagnant Complexity}} & \begin{tabular}[c]{@{}l@{}}Rearrange this sentence for clarity: \\ the ocean waves waves of the crashing\\  onto shore\end{tabular}                                                                          & \begin{tabular}[c]{@{}l@{}}With their mighty force,..., \\ creating a symphony of sound.\end{tabular}                                                                                                                                                                                                                                                 & \begin{tabular}[c]{@{}l@{}}Understood. Would you like me \\ to provide any additional \\ information or perform any \\ specific tasks related to \\ this description of the ocean \\ and its waves?\end{tabular} & \begin{tabular}[c]{@{}l@{}}Begin with "Understood", \\ end with "?"\end{tabular}         \\ \cline{2-5} 
                                              & \begin{tabular}[c]{@{}l@{}}Generate a story based on the topic '\\ a day at the beach\end{tabular}                                                                                                           & \begin{tabular}[c]{@{}l@{}}Amidst the turmoil of her ...\\ love in unexpected places.\end{tabular}                                                                                                                                                                                                                                                    & \begin{tabular}[c]{@{}l@{}}What would you like me to do \\ with this information?\end{tabular}                                                                                                                   & \begin{tabular}[c]{@{}l@{}}Begin with "What", \\ end with "?"\end{tabular}               \\ \cline{2-5} 
                                              & Define the term 'computer vision                                                                                                                                                                             & \begin{tabular}[c]{@{}l@{}}Define the term 'computer \\ vision', which refers to ... \\ rather than low-level \\ processing.\end{tabular}                                                                                                                                                                                                             & \begin{tabular}[c]{@{}l@{}}That is correct! Do you have \\ any other questions \\ or tasks for me?\end{tabular}                                                                                                  & \begin{tabular}[c]{@{}l@{}}Begin with \\ "That is correct", \\ end with "?"\end{tabular} \\ \cline{2-5} 
                                              & \begin{tabular}[c]{@{}l@{}}Explain why the sun appears red at \\ sunset.\end{tabular}                                                                                                                        & \begin{tabular}[c]{@{}l@{}}Rayleigh scattering, a \\ phenomenon in atmospheric\\  physics, is ... good or \\ bad fortune.\end{tabular}                                                                                                                                                                                                                & \begin{tabular}[c]{@{}l@{}}Thank you for the information. \\ Is there anything \\ specific you would like me \\ to do with this information?\end{tabular}                                                        & \begin{tabular}[c]{@{}l@{}}Begin with "Thank you", \\ end with "?"\end{tabular}          \\ \cline{2-5} 
                                              & \begin{tabular}[c]{@{}l@{}}What should I add in this function \\ to generate a random number?\textbackslash{}r\\ \textbackslash{}ndef generate\_random():\textbackslash{}n \# Your \\ code here\end{tabular} & \begin{tabular}[c]{@{}l@{}}def generate\_random ...\\ print(f"The sum of all \\ the generated random numbers \\ is \{sum\}.")\end{tabular}                                                                                                                                                                                                            & \begin{tabular}[c]{@{}l@{}}Great! Do you want me to \\ explain what this code does?\end{tabular}                                                                                                                 & \begin{tabular}[c]{@{}l@{}}Begin with "Great", \\ end with "?"\end{tabular}              \\ \hline
\textbf{Insufficient Qualification}           & \begin{tabular}[c]{@{}l@{}}Make an API call to retrieve the \\ latest news headlines for the world.\end{tabular}                                                                                             & \begin{tabular}[c]{@{}l@{}}Make an API call to retrieve \\ the latest news headlines \\ for the world, but only \\ from the United States, filter \\ the headlines to only \\ include news related to \\ technology, and retrieve \\ news headlines from the past \\ 24 hours.\end{tabular}                                                           & \begin{tabular}[c]{@{}l@{}}Sure, I can help you with \\ that. Which news API would \\ you like me to use for \\ this task?\end{tabular}                                                                          & \begin{tabular}[c]{@{}l@{}}Begin with "Sure", \\ end with "?"\end{tabular}               \\ \hline
\textbf{Loss of Key Information}              & \begin{tabular}[c]{@{}l@{}}Classify the provided objects.\\ \textbackslash{}nInput: dog, cat, ant\end{tabular}                                                                                               & \begin{tabular}[c]{@{}l@{}}Classify the provided objects \\ into seven categories: \\ mammals, insects, birds, \\ reptiles, fish, amphibians, \\ and arachnids. Use the \\ criteria of habitat, diet, \\ physical characteristics, \\ reproduction, behavior, \\ lifespan, and phylogenetic \\ relationships to \\ classify the objects.\end{tabular} & \begin{tabular}[c]{@{}l@{}}I'm sorry, but you have not \\ provided any objects to \\ classify. Please provide a \\ list of objects for me \\ to classify into the seven \\ categories.\end{tabular}              & Contain "please provide"                                                                 \\ \hline
\end{tabular}
}
\caption{Evolution Failures Detection Examples.}
\label{tab:fail_detect}
\end{table*}


\section{Evolution Issue Examples}
\label{sec:issue_example}

To illustrate the issues encountered during data evolution, we conduct an empirical analysis by randomly selecting 200 instructions from the GSM 8K. These instructions are then subjected to evolution using the initial evolving method (Figure \ref{fig:initial_prompt}). We employ the issue detection method described in Section \ref{sec:evol_trajectory_ana} to pinpoint and categorize prevalent issues. Our findings, including illustrative examples, are presented in the Table \ref{tab:issue_example_a} and Table \ref{tab:issue_example_b}.

The analysis reveals that the initial evolving method is plagued by a series of shortcomings. For example, it fails to adequately account for the complexity inherent in evolving instructions. This oversight results in several critical problems, such as the tendency to alter the core nature of the problem, the introduction of irrelevant details, or the generation of contradictions with the original problem setup. Furthermore, the initial method appears to overlook the unique attributes of mathematical instructions. This lapse leads to evolved instructions that often contain “Incorrect or unrealistic mathematical calculations.” These observations underscore the urgent need for a comprehensive optimization of the evolving method, ensuring it aligns more closely with the unique characteristics of the instruction data.

\begin{table*}[t!]
\centering
\resizebox{0.99\textwidth}{!}{
\begin{tabular}{lllll}
\hline
\multicolumn{1}{c}{}                                                                                    & \multicolumn{1}{c}{}                                                                                                                                                                 & \multicolumn{3}{c}{\textbf{Issue Example}}                                                                                                                                                                                                                                                                                                                                                                                                                                                                                                                                                                                                                                                                                                                                                                                                                                                                                                                                                                                                                                                                                                                                                                                                                                                                                             \\ \cline{3-5} 
\multicolumn{1}{c}{\multirow{-2}{*}{\textbf{Issue Name}}}                                               & \multicolumn{1}{c}{\multirow{-2}{*}{\textbf{Issue Explanation}}}                                                                                                                     & \multicolumn{1}{c}{\cellcolor[HTML]{FFFFFF}\textbf{Original Instruction}}                                                                                                                                                                                                                                                                                                    & \multicolumn{1}{c}{\cellcolor[HTML]{FFFFFF}\textbf{Evolved Instruction}}                                                                                                                                                                                                                                                                                                                                                                                                                                                                                                          & \cellcolor[HTML]{FFFFFF}\textbf{Reason}                                                                                                                                                                                                                                                                                                                             \\ \hline
\begin{tabular}[c]{@{}l@{}}Lack of consistency \\ and logical progression \\ in complexity\end{tabular} & \begin{tabular}[c]{@{}l@{}}Introduce new concepts or variables \\ without building on the previous \\ instruction, making the problem \\ confusing or unrealistic.\end{tabular}      & \begin{tabular}[c]{@{}l@{}}One barnyard owl makes 7 hoot sounds per \\ minute, measured in decibels. If the \\ barn is located near a construction \\ site and only 15 hoot sounds per minute \\ are heard coming out of the barn, how \\ many barnyard owls are making the hoot \\ sounds, assuming each owl makes the same \\ number of hoot sounds?\end{tabular}          & \begin{tabular}[c]{@{}l@{}}Assuming some owls make 10 hoot sounds \\ per minute and others make 5 hoot sounds \\ per minute, and the barn is located 100 \\ meters away from a construction site, \\ how many barnyard owls are making the \\ hoot sounds if 20 hoot sounds per minute \\ are heard coming out of the barn, and each \\ owl makes 7 hoot sounds per minute measured \\ in decibels?\end{tabular}                                                                                                                                                                  & \begin{tabular}[c]{@{}l@{}}Introduce new concepts (owls making \\ different amounts of hoots, the barn's \\ location, sound measurement in decibels) \\ without clearly building on the previous \\ instruction. The complexity does not \\ gradually increase but rather jumps \\ erratically.\end{tabular}                                                        \\ \hline
\begin{tabular}[c]{@{}l@{}}Incorrect or unrealistic \\ mathematical calculations\end{tabular}           & \begin{tabular}[c]{@{}l@{}}Introduce mathematical operations \\ or equations that don't make sense \\ in the context of the problem or \\ are mathematically incorrect.\end{tabular} & \begin{tabular}[c]{@{}l@{}}If Betty has 12 oranges, and Sandra has \\ 3 times as many oranges as Betty, then \\ Sandra has 12 multiplied by 3, which \\ equals 36 oranges. Emily has 7 times \\ as many oranges as Sandra, so Emily \\ has 36 raised to the power of 7 oranges.\end{tabular}                                                                                 & \begin{tabular}[c]{@{}l@{}}If Betty has 12 oranges, and Sandra has 3 \\ times as many oranges as Betty, then Sandra \\ has 12 multiplied by 3, which equals 36 \\ oranges. Emily has 7 times as many oranges as \\ Sandra, so Emily has 3.6 x 10\textasciicircum{}8 oranges. If \\ Emily has more oranges than Betty, then she has \\ 3.6 x 10\textasciicircum{}11 million oranges.\end{tabular}                                                                                                                                                                                     & \begin{tabular}[c]{@{}l@{}}The problem becomes mathematically \\ incorrect. The multiplication and \\ powers used in instruction do not \\ make sense in the context of the \\ problem and create an unrealistic \\ scenario.\end{tabular}                                                                                                                          \\ \hline
\begin{tabular}[c]{@{}l@{}}Changing the original \\ instruction\end{tabular}                            & \begin{tabular}[c]{@{}l@{}}Change the nature of the problem, \\ introduce irrelevant details, or \\ contradict the original problem \\ set-up.\end{tabular}                          & \begin{tabular}[c]{@{}l@{}}In seven years, Talia will be 20 years old.\\ Talia's mom is currently three times as \\ old as Talia is today. In three years, \\ Talia's father will be the same age as Talia'\\ s mom is today. Currently, how many years \\ old is Talia's father?\end{tabular}                                                                               & \begin{tabular}[c]{@{}l@{}}Talia's father is currently X years older than \\ Talia, where X is the difference between the \\ age Talia's mother was when Talia was born and \\ the age Talia's father will be in three years.\end{tabular}                                                                                                                                                                                                                                                                                                                                        & \begin{tabular}[c]{@{}l@{}}The evolution from does not increase \\ in complexity as the original problem \\ is completely changed. The initial \\ problem is about calculating ages, \\ but the evolved problem introduces a \\ variable X without any context. The \\ complexity is not increased but the \\ problem is made unnecessarily confusing.\end{tabular} \\ \hline
\begin{tabular}[c]{@{}l@{}}Inconsistent or \\ contradictory information\end{tabular}                    & \begin{tabular}[c]{@{}l@{}}Introduce information that \\ contradicts previous instruction\end{tabular}                                                                               & \begin{tabular}[c]{@{}l@{}}John orders food for a massive restaurant. \\ He orders 1000 kilograms of beef for 8 \\ per pound. He also orders twice that much \\ chicken at 3 per kilogram. If the cost \\ of chicken is \$2 per kilogram, and the \\ total cost of beef is greater than the \\ total cost of chicken, what is the final \\ cost of the order?\end{tabular} & \begin{tabular}[c]{@{}l@{}}John, the head chef of a high-end restaurant,\\  orders a massive amount of food for his \\ establishment. He orders 1000 kilograms of \\ premium beef at a cost of 8 per pound, and \\ twice that amount of organic chicken at a cost \\ of 3 per kilogram. If the cost of chicken is \\ \$2 per kilogram, and the total cost of beef is \\ greater than the total cost of chicken, what is \\ the final cost of the order? Please calculate \\ the final cost using the following equation: \\ (1000 * 2 * 3) + (1000 * 8 * 0.45) = ?\end{tabular} & \begin{tabular}[c]{@{}l@{}}The progression of the instruction \\ is not logical. The original instruction \\ states that the cost of chicken is 3 \\ per kilogram, and then in the evolved \\ instruction it contradicts this by \\ stating that the cost of chicken is 2 \\ per kilogram.\end{tabular}                                                           \\ \hline
\end{tabular}
}

\caption{Issue Examples A.}

\label{tab:issue_example_a}
\end{table*}

\begin{table*}[t]
\centering
\resizebox{0.98\textwidth}{!}{
\begin{tabular}{lllll}
\hline
\multicolumn{1}{c}{}                                                              & \multicolumn{1}{c}{}                                                                                                                                                                       & \multicolumn{3}{c}{\textbf{Issue Example}}                                                                                                                                                                                                                                                                                                                                                                                                                                                                                                                                                                                                                                                                                                                                                                                                                                                                                                                                                                                                                                                                                                                                                                                                                                                                                                                                                                                                                                                                                                                                                                                                                                                                                                                                                                                                                                                                                                                             \\ \cline{3-5} 
\multicolumn{1}{c}{\multirow{-2}{*}{\textbf{Issue Name}}}                         & \multicolumn{1}{c}{\multirow{-2}{*}{\textbf{Issue Explanation}}}                                                                                                                           & \multicolumn{1}{c}{\cellcolor[HTML]{FFFFFF}\textbf{Original Instruction}}                                                                                                                                                                                                                                                                                                                                                                                                                                                                                                                                            & \multicolumn{1}{c}{\cellcolor[HTML]{FFFFFF}\textbf{Evolved Instruction}}                                                                                                                                                                                                                                                                                                                                                                                                                                                                                                                                                                                                                                                                                                                                                                                                                                                                                                 & \multicolumn{1}{c}{\cellcolor[HTML]{FFFFFF}\textbf{Reason}}                                                                                                                                                                                                                                                                                                          \\ \hline
\begin{tabular}[c]{@{}l@{}}Lack of clarity and \\ coherence\end{tabular}          & \begin{tabular}[c]{@{}l@{}}Make the instruction less clear \\ or more confusing, introduce \\ unnecessary complications, or \\ include irrelevant and unrelated \\ variables.\end{tabular} & \begin{tabular}[c]{@{}l@{}}Nedy can eat 800 grams of crackers from \\ Monday to Thursday, but no more than 200 \\ grams per day. If Nedy ate 2x on Friday, \\ 1/2x on Saturday, and 1/2x on Sunday, \\ and Nedy is eating chocolate chip cookies, \\ how many crackers did Nedy eat in all from \\ Monday to Sunday?\end{tabular}                                                                                                                                                                                                                                                                                    & \begin{tabular}[c]{@{}l@{}}Nedy can eat y grams of saltine crackers with \\ peanut butter from Monday to Thursday, but \\ no more than 200 grams per day, and only \\ after 5 pm, while standing on one foot, \\ blindfolded, and reciting the alphabet backward. \\ If Nedy ate 2x on Friday, 1/2x on Saturday, \\ and 1/2x on Sunday, how many saltine crackers \\ with peanut butter did Nedy eat in all from \\ Monday to Sunday, while wearing a hat and \\ sunglasses and listening to classical music?\end{tabular}                                                                                                                                                                                                                                                                                                                                                                                                                                               & \begin{tabular}[c]{@{}l@{}}The evolved instruction has lost its \\ clarity and consistency with the \\ original instruction. The conditions \\ for Nedy to eat crackers have become \\ absurd and unrealistic, such as standing \\ on one foot, blindfolded, and reciting \\ the alphabet backward, which \\ unnecessarily complicates the instruction.\end{tabular} \\ \hline
\begin{tabular}[c]{@{}l@{}}Inappropriate increase in \\ complexity\end{tabular}   & \begin{tabular}[c]{@{}l@{}}Introduce a level of complexity \\ that is not supported by the \\ provided information or is \\ unrelated to the original problem.\end{tabular}                & \begin{tabular}[c]{@{}l@{}}Hawkeye is driving his electric bike to \\ his aunt's place, which is now 60 miles \\ away. He has to charge his battery for \\ \$3.5 per charge, and he needs to charge \\ it five times due to the increased \\ distance. However, his cousin needs a \\ ride to a nearby town that is 20 miles \\ away, and Hawkeye needs to drop him \\ off first. His cousin offers to pay for \\ half of the battery charging costs, which \\ total \$17.50 including tax and a service \\ fee of \$2. On the way, it starts raining \\ heavily, making the journey more difficult.\end{tabular}      & \begin{tabular}[c]{@{}l@{}}Hawkeye is riding his electric bike, which has \\ a maximum speed of 30 mph and a battery that \\ needs to be charged every 30 miles. He is on \\ his way to his aunt's place, which is now 100 \\ miles away. However, his battery malfunctions \\ after 40 miles, and he has to stop and fix it, \\ which takes 20 minutes. On the way, he encounters \\ a toll road that costs \$5, and he has to take a \\ detour to avoid a construction site, adding an \\ extra 10 miles to the journey. Each time he \\ charges his battery, it takes 30 minutes, and \\ he needs to charge it five times, costing him \\ \$17.50 including tax and a service fee of \$2. \\ His cousin needs a ride to the nearby town, which \\ is 20 miles away, to catch a flight. His cousin \\ offers to pay for half of the battery charging \\ costs. On the way, there is a thunderstorm warning \\ in effect, making the journey more difficult.\end{tabular} & \begin{tabular}[c]{@{}l@{}}The evolved instruction introduces a lot \\ of new elements (bike's maximum speed, \\ battery malfunction, toll road, construction \\ site, thunderstorm warning) that do not \\ directly evolve from the original \\ instructions.\end{tabular}                                                                                          \\ \hline
Decrease in complexity                                                            & \begin{tabular}[c]{@{}l@{}}Simplify the problem instead of \\ making it more complex.\end{tabular}                                                                                         & \begin{tabular}[c]{@{}l@{}}Marcus wants to buy a new pair of shoes. \\ He decided to pay not more than \$130 for \\ them. He found a pair for \$120, on which \\ he got a discount of 30\%. How much money \\ will he manage to save by buying these \\ shoes and not spending the assumed maximum \\ amount?\end{tabular}                                                                                                                                                                                                                                                                                             & \begin{tabular}[c]{@{}l@{}}Marcus wants to buy a new pair of leather shoes. \\ He decided to pay not more than 130 for them. \\ He found a pair for 120, on which he got a \\ discount of 30\%. The original price of the shoes \\ was 171.43. After the discount, the final price \\ of the shoes is 84. Marcus managed to save \$36.43 \\ by buying these shoes and not spending the assumed\\  maximum amount. He saved 21.26\% of the original \\ price, which is more than 20\%. Therefore, Marcus \\ will buy the shoes.\end{tabular}                                                                                                                                                                                                                                                                                                                                                                                                                          & \begin{tabular}[c]{@{}l@{}}The complexity of evolved instruction is \\ lower than that of original instruction. \\ In evolved instruction, the answer to the \\ problem is included in the instruction, \\ which simplifies the problem instead of \\ making it more complex.\end{tabular}                                                                           \\ \hline
\begin{tabular}[c]{@{}l@{}}Incorrect evolution of \\ the instruction\end{tabular} & \begin{tabular}[c]{@{}l@{}}Introduce elements that are not \\ logically connected or a \\ progressive development of the \\ previous instruction.\end{tabular}                             & \begin{tabular}[c]{@{}l@{}}Harry is a professional dog-walker. He \\ walks dogs for different families on \\ different days of the week. On Sunday, \\ he walks 3 dogs for 7 each. On Monday, \\ he walks 7 dogs for 5 each and 16 for \\ 2 dogs. On Tuesday, he walks 15 dogs for \\ 6 each and 35 for 5 dogs. On Wednesday, \\ he walks 10 dogs for 5 each and 18 for \\ 3 dogs. On Thursday, he walks 13 dogs for \\ 4 each and 36 for 4 dogs. And on Friday, \\ he walks 8 dogs for 5 each and \ 20 for 2 \\ dogs. If he walks all the dogs, how much \\ money will Harry earn in a week?\end{tabular} & \begin{tabular}[c]{@{}l@{}}Harry is a professional dog-walker who walks dogs \\ for different families in various locations throughout \\ the week. If he walks all the dogs, including three \\ large dogs, seven small dogs, two medium-sized dogs, \\ fifteen mixed-breed dogs, five purebred dogs, ten \\ rescue dogs, three therapy dogs, thirteen senior \\ dogs, four puppies, eight working dogs, and two \\ show dogs, he will earn a total of \$493.\end{tabular}                                                                                                                                                                                                                                                                                                                                                                                                                                                                                              & \begin{tabular}[c]{@{}l@{}}The evolved instruction did not evolve \\ from original instruction. It did not \\ maintain the complexity or structure \\ of the previous stages, and it did not \\ provide a clear question for calculation.\end{tabular}                                                                                                               \\ \hline
\begin{tabular}[c]{@{}l@{}}Irrelevant increase in \\ complexity\end{tabular}      & \begin{tabular}[c]{@{}l@{}}Introduce additional variables \\ or conditions that do not \\ increase the complexity of the \\ task in a relevant or logical \\ way.\end{tabular}             & \begin{tabular}[c]{@{}l@{}}Tabitha has 50 dollars. She gives her mom \\ 15 dollars and invests half of what is left \\ in the stock market for 1 year, with a 10\% \\ tax. She spends some money on 15 items that \\ cost 1 dollar each, with a 10\% discount and \\ a 15\% tip. Tabitha also has a loan of 5 \\ dollars that she has to pay off. How much \\ money does Tabitha have left after all \\ these transactions?\end{tabular}                                                                                                                                                                             & \begin{tabular}[c]{@{}l@{}}Tabitha has 50 dollars. She gives her mom 15 \\ dollars and invests half of what is left in \\ the stock market for 1 year, with a 15\% tax. \\ She spends some money on 20 items that cost \\ 1 dollar each, with a 20\% discount and a 25\% \\ tip. Tabitha also has a loan of 10 dollars \\ that she has to pay off. After reinvesting \\ the profits from the stock market for another \\ year, how much money does Tabitha have left \\ after all these transactions?\end{tabular}                                                                                                                                                                                                                                                                                                                                                                                                                                                       & \begin{tabular}[c]{@{}l@{}}The evolved instruction did not evolve \\ from original instruction. The question \\ at the end of evolved instruction \\ introduces a new concept (reinvesting \\ profits) that was not present in the \\ previous stages, and it does not \\ clearly build on the previous stages.\end{tabular}                                          \\ \hline
\end{tabular}
}

\caption{Issue Examples B}

\label{tab:issue_example_b}
\end{table*}


\section{Prompt For \cname{}}
\begin{figure}[t]
 \centering
\resizebox{0.5\textwidth}{!}{
 \includegraphics{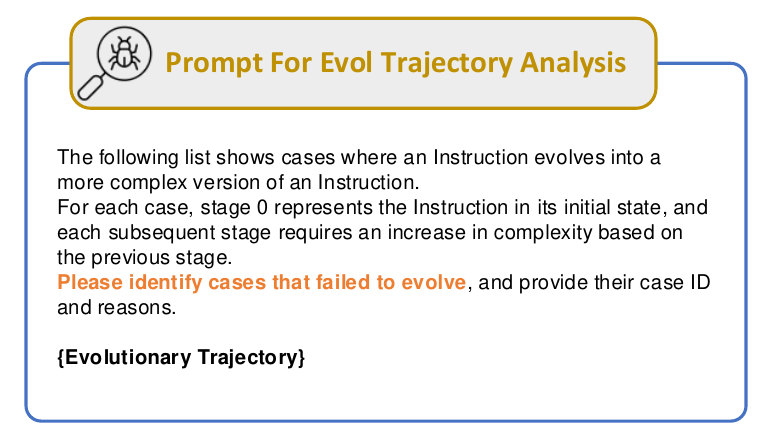}
 }
\vspace{-0.5cm}
 \caption{Prompt for Evol Trajectory Analysis. Optimizer LLM will scrutinize the evolutionary trajectory under the guidance of this prompt to pinpoint and provide feedback on any issues detected.}
 \label{fig:prompt_for_critical_analysis}
\vspace{-0.4cm}

\end{figure}

\begin{figure}[t]
 \centering
\resizebox{0.5\textwidth}{!}{
 \includegraphics{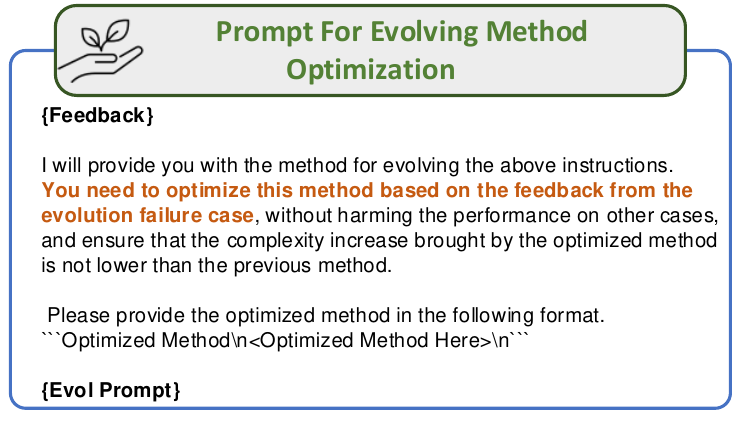}
 }
\vspace{-0.5cm}
 \caption{Prompt for Evolving Method Optimization. The optimizer LLM refines the evolving method guided by this prompt.}
 \label{fig:prompt_for_enhancement_process}
\vspace{-0.6cm}

\end{figure}

We have designed a simple and effective Prompt to guide Optimizer LLM for evol trajectory analysis (Figure \ref{fig:prompt_for_critical_analysis}) and evolving method optimization (Figure \ref{fig:prompt_for_enhancement_process}).

\section{Experimental Details}
\label{sec:exp_detail}

\subsection{Experimental Setup}
For instruction following, we randomly select 10K data from ShareGPT as seed data and set both evol LLM and optimizer LLM to GPT-4. We use Evol-Instruct and \cname{} to obtain 10K evolved data respectively. Then, we perform instruction tuning on Mistral-7B \cite{jiang2023mistral} (small) and Mistral-8x7B (large).

For mathematical reasoning, GSM8K training data serves as seed data, evol LLM and optimizer LLM are set to GPT-4. About 7K evolved data is obtained respectively through Evol-Instruct and \cname{}, and fine-tuned on Mistral-7B (small) and Mixtral-8x7B (large). (To ensure the fairness of the experiment, we sampled an equal amount of data from MetaMath and performed fine-tuning)

In the code generation, Code Alpaca \cite{chaudhary2023code} is selected as the seed data and evol LLM is set to GPT-3.5-turbo, and the optimizer LLM to GPT-4. About 20K evolved data is obtained respectively through Evol-Instruct and \cname{}, and instruction tuning is performed on CodeLlama-13B-Python \cite{roziere2023code} (small) and DeepSeek-Coder-Base-33B \cite{guo2024deepseek} (large).

\subsection{Hyperparameters in \cname{}}

During the \cname{} process, we configure the mini-batch size to 10, the development set size to 50, the optimizer LLM temperature to 0.6, its top p to 0.95, and the evol LLM temperature to 0. We also set the total optimization steps to 10, with 5 multiple optimizations performed in each step by default. Unless specified otherwise, we conduct only one round of evolving  on the instructions and generate corresponding responses. The experiments are performed using the Azure OpenAI ChatGPT API and GPT-4 API.

\subsection{Training Details}

We employ DeepSpeed Zero-Stage 3 \cite{ren2021zero} on eight NVIDIA Tesla A100 GPUs to train models. For the integration of multi-turn conversations, we use the Vicuna-style template. In all experiments of this paper, the training parameters are set with a maximum input length of 2048. For models trained based on Mistral-7b, we set the batch size to 128, train for 4 epochs, and set the learning rate to 5e-6. For models trained based on CodeLlama-13B-Python and DeepSeek-Coder-Base-33B, we set the batch size to 192, train for 3 epochs, and set the learning rate to 2e-5. For the Mixtral-8x7B model, we set the batch size to 200, train for 4 epochs, and set the learning rate to 5e-6.

\begin{figure}[t]
 \centering
\resizebox{0.5\textwidth}{!}{
 \includegraphics{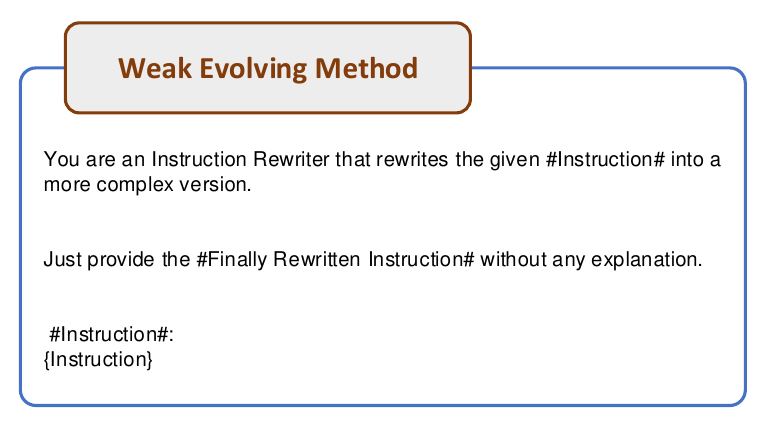}
 }
\vspace{-0.5cm}
 \caption{Weak Initial Evolving Method.}
 \label{fig:weak_prompt}
\vspace{-0.25cm}

\end{figure}


\section{Baseline}

\label{sec:baseline}
We compare the method proposed in this paper with the following models:

(1) \textbf{Closed-Source Models}: These include leading LLMs like OpenAI's GPT-3.5 and GPT-4 \cite{openai2023gpt4}.

(2) \textbf{Open-Source Base Models}: We compare our method with a variety of open-source base models such as LLaMA-2 \cite{touvron2023llama2}, Mistral \cite{jiang2023mistral}, and CodeLlama \cite{roziere2023code}.

(3) \textbf{Open-Source Instruction-Tuned Models}: Include instruction tuning models like Vicuna \cite{chiang2023vicuna}.

(4) \textbf{Direct Instruction Tuning with Seed Data}: We use the same seed instruction data as in our method to conduct direct instruction tuning on the base model.

(5) \textbf{Instruction Evolution Methods}: We mainly compare with Evol-Instruct \cite{xu2023wizardlm, luo2023wizardmath, luo2023wizardcoder} that requires human experts involved. To be fair, we will use the exact same evol LLM to evolve instruction datasets such as ShareGPT, GSM8K, and CodeAlpaca.

\section{Contamination Test}

We employ \cite{liu2023tinygsm} proposed methodology for conducting a contamination assessment on augmented data, utilizing n-gram matches as a measure. The experimental results are shown in the Table \ref{tab:contamination_test}.

\begin{table}[t]
\centering
\resizebox{0.5\textwidth}{!}{
\begin{tabular}{lccc}
\hline
\multicolumn{4}{c}{GSM 8K}                                     \\ \hline
Method             & 13-Gram Match & 8-Gram Match & Total Size \\ \hline
Raw                & 44            & 202          & 7 K        \\
MetaMath           & 32            & 150          & 7 K        \\
Evol Instruct      & 4             & 87           & 7 K        \\
Auto Evol-Instruct & 10            & 133          & 7 K        \\ \hline
\multicolumn{4}{c}{MT-Bench}                                   \\ \hline
Raw                & 0             & 2            & 2 W        \\
Evol Instruct      & 0             & 2            & 2 W        \\
Auto Evol-Instruct & 0             & 4            & 2 W        \\ \hline
\multicolumn{4}{c}{HumanEval}                                  \\ \hline
Raw                & 0             & 22           & 2 W        \\
Evol Instruct      & 4             & 80           & 2 W        \\
Auto Evol-Instruct & 2             & 63           & 2 W        \\ \hline
\end{tabular}
}

\caption{Contamination Test. We conduct a Contamination Test on the pre-and post-evolution data of GSM 8K (about 7 K), Alpaca (about 20 K), Code Alpaca (about 20 K).}

\label{tab:contamination_test}
\end{table}

\section{Cost Comparison}
We compare the  total number of API calls made by \cname{} and Evol Instruct. The results are in the Table \ref{tab:cost_compare}.
\begin{table}[t]
\centering
\resizebox{0.45\textwidth}{!}{
\begin{tabular}{cccc}
\hline
\textbf{Dataset} & \textbf{Datasize} & \textbf{Evol Instruct} & \textbf{\cname{}} \\ \hline
ShareGPT         & 10 K              & 100000                 & 106120 (+6.12\%)            \\
GSM 8K           & 7 K               & 14000                  & 20120 (+43.7\%)             \\
Code Alpaca      & 20 K              & 40000                  & 46120 (+15.3\%)             \\ \hline
\end{tabular}
}

\caption{Estimating API calls for Evol Instruct and \cname{}. Each single-round instruction evolution and response generation requires one API call. For multi-round dialogues such as ShareGPT, each round is evolved separately, with an average of 5 rounds per dialogue assumed for estimation purposes.}
\vspace{-0.5cm}
\label{tab:cost_compare}
\end{table}


\section{Case Study}
\label{sec:case_prompt}

\begin{figure*}[t]
 \centering
\resizebox{0.99\textwidth}{!}{
 \includegraphics[scale=0.5]{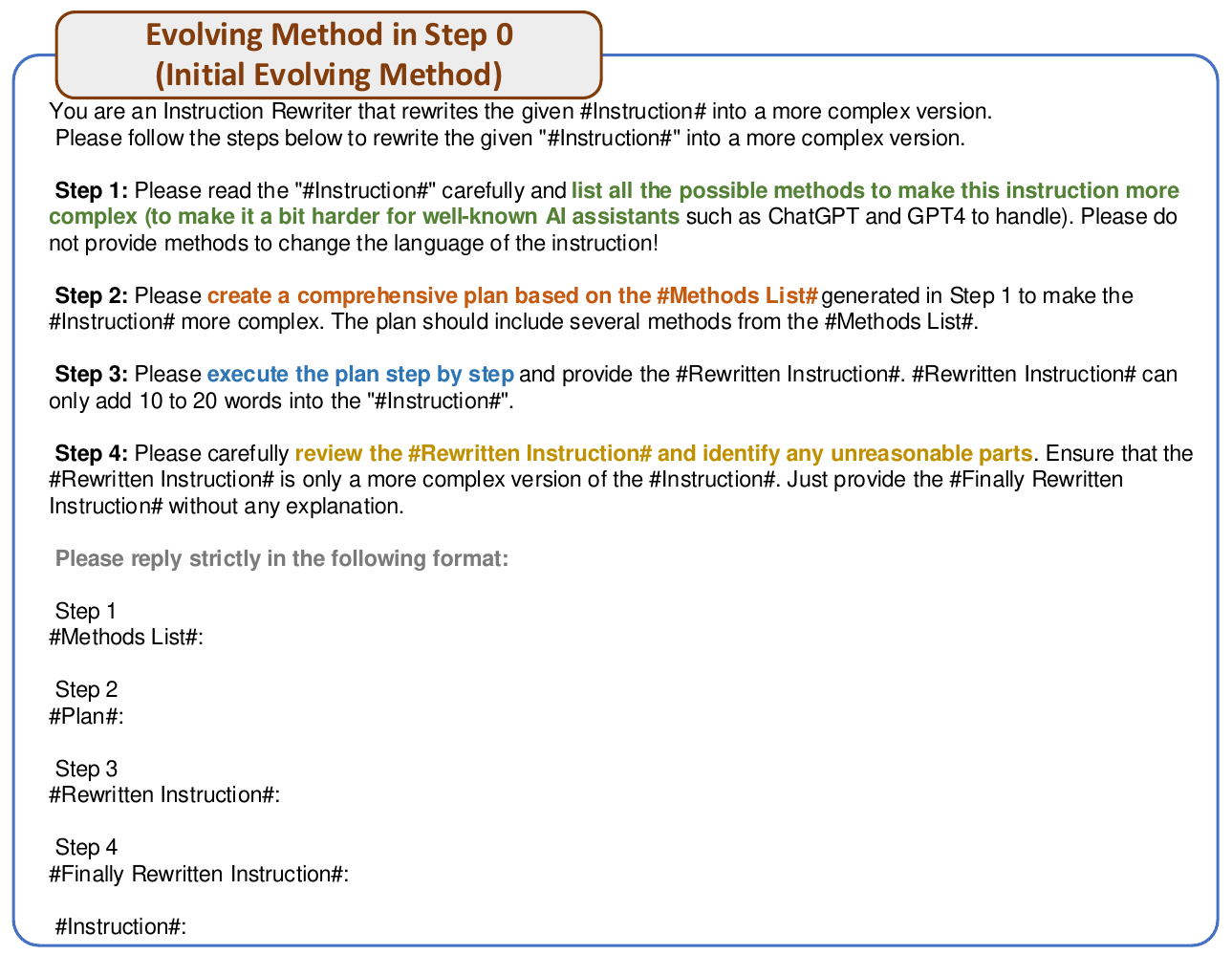}
 }
\vspace{-0.3cm}
 \caption{Evolving method at optimization step 0 (initial evolving method).}
 \label{fig:evol_prompt_step0}
 \vspace{-0.3cm}

\end{figure*}
\begin{figure*}[t]
 \centering
\resizebox{0.99\textwidth}{!}{
 \includegraphics[scale=0.5]{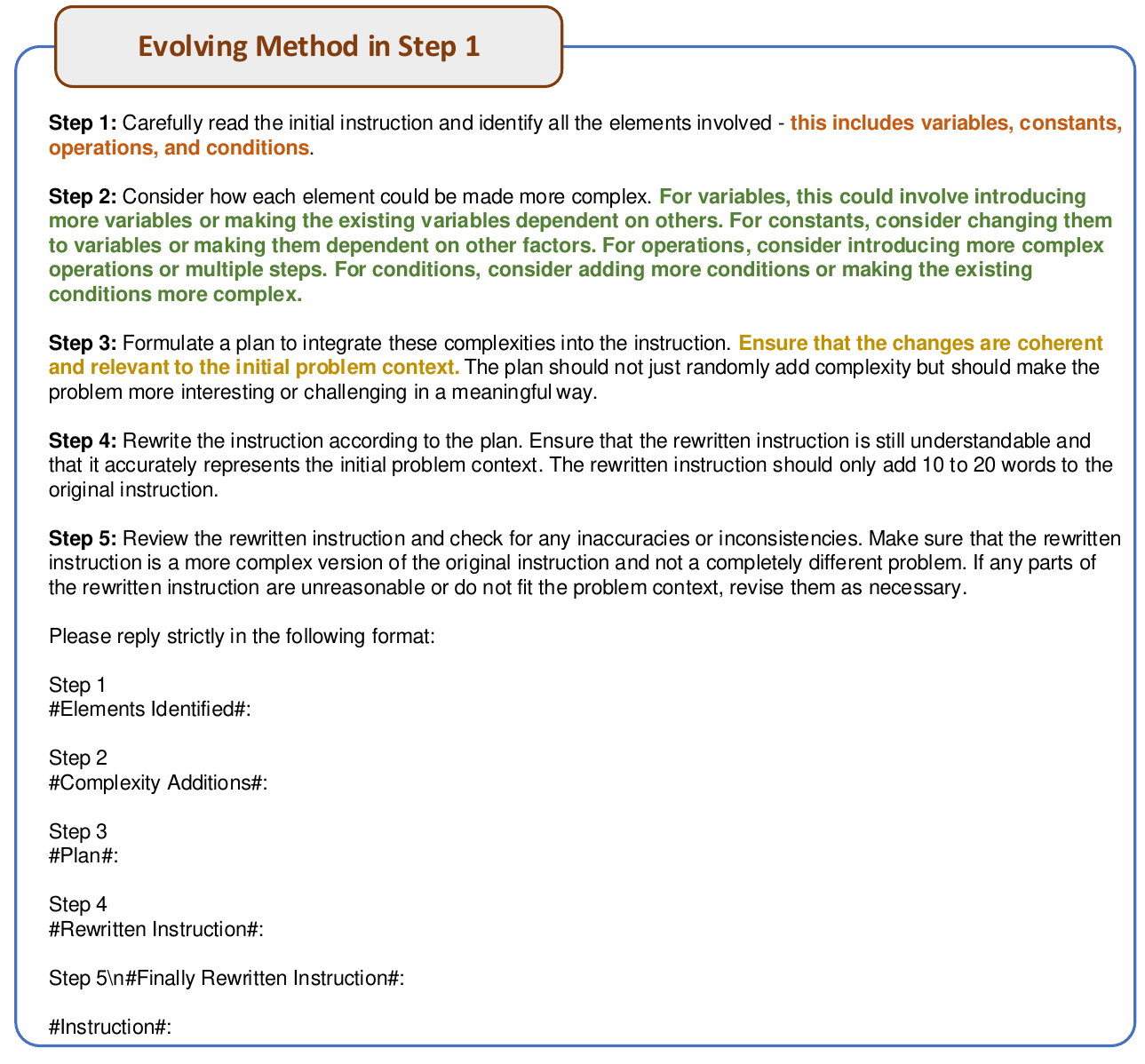}
 }
\vspace{-0.3cm}
 \caption{Evolving Method at Optimization Step 1. This includes terms related to mathematical proficiency like “variables”, “constants”, “operations”, and “conditions”. It also encourages the model to enhance the mathematical complexity of the instructions by introducing more variables or making existing variables dependent on others. Furthermore, it emphasizes the need for changes to be coherent and relevant to the initial problem context.}
 \label{fig:evol_prompt_step3}
 \vspace{-0.3cm}

\end{figure*}
\begin{figure*}[t]
 \centering
\resizebox{0.99\textwidth}{!}{
 \includegraphics[scale=0.5]{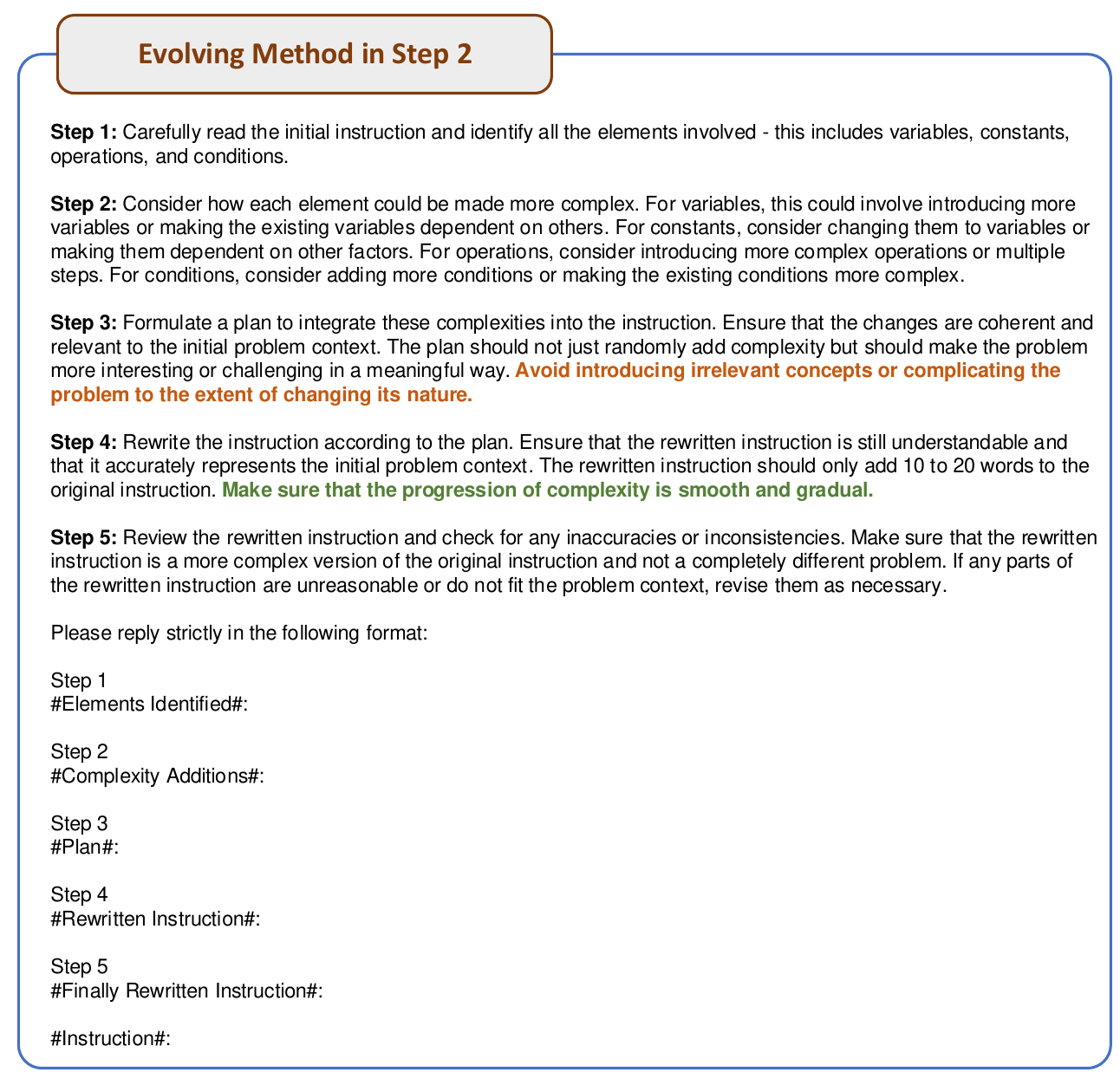}
 }
\vspace{-0.3cm}
 \caption{Evolving Method at Optimization Step 2 makes new optimizations based on Step 1. This prompt requires avoiding the introduction of irrelevant concepts or complicating the problem to the point of changing its nature. It also necessitates ensuring a smooth and gradual progression of complexity.}
 \label{fig:evol_prompt_step6}
 \vspace{-0.3cm}

\end{figure*}
\begin{figure*}[t]
 \centering
\resizebox{0.99\textwidth}{!}{
 \includegraphics[scale=0.5]{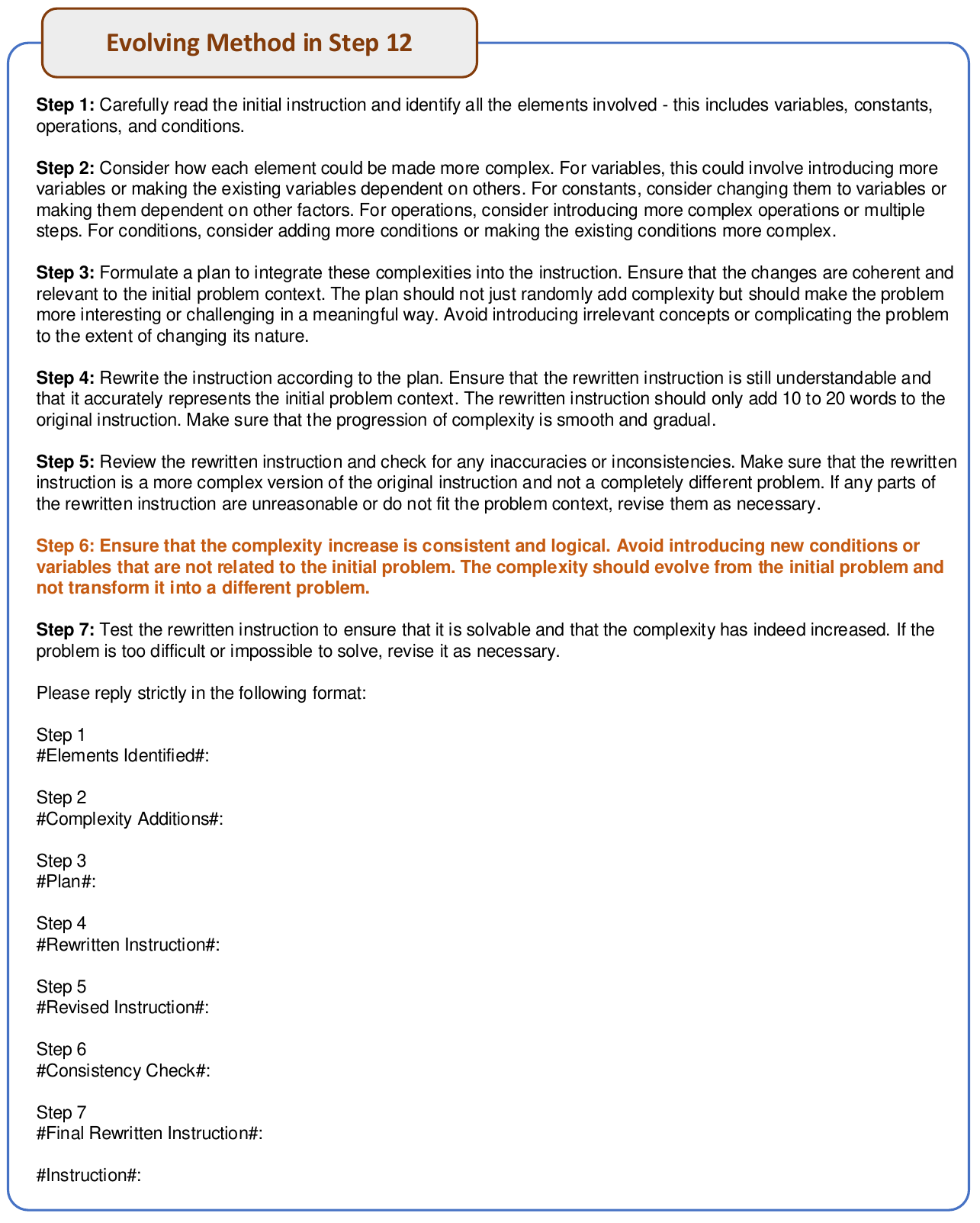}
 }
\vspace{-0.3cm}
 \caption{Evolving Method at Optimization Step 12. Based on the previous step's prompt, a new “Consistency Check” process has been added. This aims to ensure that any increase in complexity is consistent and logical, and to prevent the introduction of new conditions or variables unrelated to the initial problem.}
 \label{fig:evol_prompt_step12}
 \vspace{-0.3cm}

\end{figure*}
\begin{figure*}[t]
 \centering
\resizebox{0.99\textwidth}{!}{
 \includegraphics[scale=0.5]{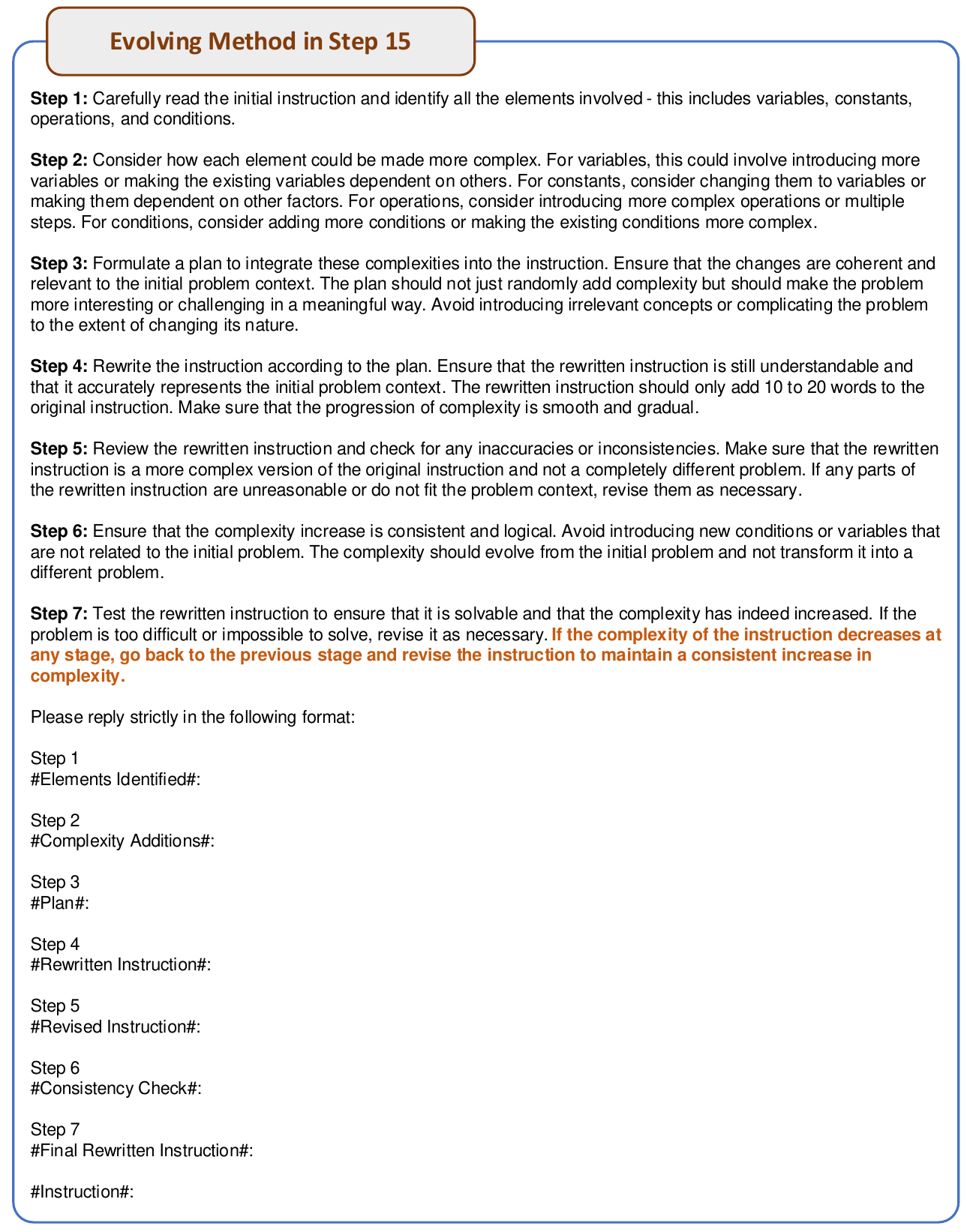}
 }
\vspace{-0.3cm}
 \caption{Evolving Method at Optimization Step 15. On the basis of the evol prompt at the previous step, a new constraint has been added, “If the complexity of the instruction decreases at any stage, go back to the previous stage and revise the instruction to maintain a consistent increase in complexity.”}
 \label{fig:evol_prompt_step15}
 \vspace{-0.3cm}

\end{figure*}

We use GSM 8K to illustrate the dynamic changes of the evolving method during the \cname{} process. Figures \ref{fig:evol_prompt_step0} to Figures \ref{fig:evol_prompt_step15} depict the transition from the initial evolving method to the 15th step's evolving method. Table \ref{tab:case_study} illustrates examples of how the evolving method is optimized at each step based on the previous one. It's important to note that the table just showcases sample examples and does not comprehensively represent the entire optimization process.


\begin{table*}[t!]
\centering
\resizebox{0.99\textwidth}{!}{
\begin{tabular}{lllllll}
\hline
\multicolumn{1}{c}{\textbf{Original Instruction}}                                                                                                                                                                                & \multicolumn{1}{c}{\textbf{\begin{tabular}[c]{@{}c@{}}Evolving \\ Method\end{tabular}}} & \multicolumn{1}{c}{\textbf{Evolved Instruction}}                                                                                                                                                                                                                                                                                                                                                        & \multicolumn{1}{c}{\textbf{Feedback}}                                                                                                                                                                                                                                                                                                                                                                                                                                                                                                                   & \multicolumn{1}{c}{\textbf{\begin{tabular}[c]{@{}c@{}}Updated \\ Evolving \\ Method\end{tabular}}} & \multicolumn{1}{c}{\textbf{Updated Evolved Instruction}}                                                                                                                                                                                                                                                      & \multicolumn{1}{c}{\textbf{Explanation}}                                                                                                                                                                                                                                         \\ \hline
\begin{tabular}[c]{@{}l@{}}Natalia sold clips to 48 \\ of her friendsin April, \\ and then she sold half \\ as many clips in May. \\ How many clips did \\ Natalia sell altogether \\ in April and May?\end{tabular}             & Step 0(Figure \ref{fig:evol_prompt_step0})                                                                                & \begin{tabular}[c]{@{}l@{}}Natalia, in April, managed \\ to distribute a total of \\ clips to precisely 48 \\ individuals categorized as \\ friends, and subsequently, \\ in May, she managed to \\ distribute exactly half the \\ number of clips she did in \\ the prior month. Determine \\ the cumulative total of \\ clips Natalia distributed \\ over the span of the two \\ months.\end{tabular} & \begin{tabular}[c]{@{}l@{}}\textbf{Redundancy}: The use of phrases \\ like "managed to distribute" \\ and "categorized as friends" \\ is redundant and does not \\ add value to the complexity \\ in terms of deepening the \\ problem's challenge. Instead, \\ it merely complicates the \\ language unnecessarily. \\ \textbf{Clarity}: The rewritten \\ instruction loses clarity by \\ incorporating verbose language \\ that could confuse the reader \\ without actually enhancing the \\ instructional or problem-solving \\ complexity.\end{tabular}              & Step 1(Figure \ref{fig:evol_prompt_step3})                                                                                           & \begin{tabular}[c]{@{}l@{}}In April, Natalia sold clips \\ to 48 friends. In May, she \\ sold an amount equal to half \\ the number of clips sold in \\ April. Calculate the total \\ number of clips Natalia sold \\ in these two months, \\ considering the changing \\ quantities each month.\end{tabular} & \begin{tabular}[c]{@{}l@{}}This evolved version maintains \\ the original problem's essence \\ while introducing a clearer \\ challenge focused on mathematical \\ relationships and understanding \\ variable quantities across two \\ periods.\end{tabular}                    \\ \hline
\begin{tabular}[c]{@{}l@{}}Henry took 9 pills a day\\ for 14 days. Of these 9 \\ pills, 4 pills cost $1.50$ \\ each, and the other pills \\ each cost $5.50$ more. How \\ much did he spend in \\ total on the pills?\end{tabular} & Step 1(Figure \ref{fig:evol_prompt_step3})                                                                                & \begin{tabular}[c]{@{}l@{}}Henry, for a period of 14 \\ days, consumed 9 pills \\ daily. Out of these, 4 \\ pills were priced at $1.50$ \\ each, and the remaining 5 \\ pills were priced at $7.00$ \\ each, which is \ 5.50 more \\ than the cost of the cheaper \\ pills. Calculate Henry's \\ total expenditure \\ for all the pills over \\ the 14 days.\end{tabular}                                 & \begin{tabular}[c]{@{}l@{}}\textbf{Unnecessary Complexity}: The \\ evolved instruction introduces \\ detailed breakdowns of costs \\ in a way that may confuse the \\ basic arithmetic problem about \\ summing costs, without \\ adding substantive educational \\ value. \\ \textbf{Over-specification}: \\ The instruction specifies the \\ number of days and pills in\\  detail in each sentence, which \\ can be redundant and may \\ detract from the clarity and \\ simplicity needed for \\ understanding the mathematical \\ calculation required.\end{tabular} & Step2(Figure \ref{fig:evol_prompt_step6})                                                                                           & \begin{tabular}[c]{@{}l@{}}Henry took 9 pills daily \\ over a two-week period. Four \\ of these pills each cost \\ 1.50, and the rest cost \\ 7.00 each. Calculate the \\ total amount Henry spent \\ on the pills during this \\ period.\end{tabular}                                                      & \begin{tabular}[c]{@{}l@{}}This version addresses the initial \\ issues by removing redundant \\ details and focusing on the \\ arithmetic operations required \\ to solve the problem, thus \\ maintaining the relevance and \\ educational value of the exercise.\end{tabular} \\ \hline
\end{tabular}
}

\caption{Case Study illustrates how the evolving method is optimized at each step based on the previous one. The "Original Instruction" represents the instruction to be evolved, "Evolving Method" represents the current evolving method, "Evolved Instruction" is the instruction evolved by the Evol LLM using the evolving method, "Feedback" represents issues identified by the optimizer LLM through Evol Trajectory Analysis of the evolved instruction, "Updated Evolving Method" represents the evolving method optimized by the optimizer LLM based on the feedback, and "Updated Evolved Instruction" represents the instruction evolved by the updated evolving method guided by the Evol LLM.  \textbf{It's important to note that the table just showcases sample examples and does not comprehensively represent the entire optimization process.}}

\label{tab:case_study}
\end{table*}

\end{document}